\documentclass[10pt,twocolumn,letterpaper]{article}

\usepackage{iccv}
\usepackage{times}
\usepackage{epsfig}
\usepackage{graphicx}
\usepackage{amsmath}
\usepackage{amssymb}
\usepackage{cancel}
\usepackage[dvipsnames]{xcolor}
\usepackage{multirow}
\usepackage{colortbl}

\usepackage{capt-of}

\usepackage{algorithm}
\usepackage{algorithmic}
\usepackage[accsupp]{axessibility}
\usepackage{multirow}
\usepackage{adjustbox}
\usepackage{bbm}

\usepackage{tablefootnote}

\usepackage[pagebackref=true,breaklinks=true,letterpaper=true,colorlinks,bookmarks=false]{hyperref}

\iccvfinalcopy 

\def\iccvPaperID{8602} 

\ificcvfinal\fi
\newcommand{\thatis}{\textit{i.e.}}
\newcommand{\withrespectto}{\textit{w.r.t.}}

\newcommand{\titlecolor}[1]{\textcolor{black}{#1}}

\begin{document}

\title{\titlecolor{MAMo:} Leveraging \titlecolor{\underline{\textcolor{black}{M}}}emory and \titlecolor{\underline{\textcolor{black}{A}}}ttention for \\ \titlecolor{\underline{\textcolor{black}{Mo}}}nocular Video Depth Estimation \vspace{-10pt}}

\author{Rajeev Yasarla, Hong Cai, Jisoo Jeong, Yunxiao Shi, Risheek Garrepalli, and Fatih Porikli \\
Qualcomm AI Research\thanks{Qualcomm AI Research is an initiative of Qualcomm Technologies, Inc.}\\
{\tt\small \{ryasarla, hongcai, jisojeon, yunxshi, rgarrepa,  fporikli\}@qti.qualcomm.com}
}

\maketitle
\ificcvfinal\thispagestyle{empty}\fi

\setlength{\belowdisplayskip}{4pt} 
\setlength{\abovedisplayskip}{4pt} 

\begin{abstract} 
\vspace{-8pt}
We propose MAMo, a novel memory and attention framework for monocular video depth estimation. MAMo can augment and improve  any single-image depth estimation networks into video depth estimation models, enabling them to take advantage of the temporal information to predict more accurate depth. In MAMo, we augment model with memory which aids the depth prediction as the model streams through the video. Specifically, the memory stores learned visual and displacement tokens of the previous time instances. This allows the depth network to cross-reference relevant features from the past when predicting depth on the current frame. We introduce a novel scheme to continuously update the memory, optimizing it to keep tokens that correspond with both the past and the present visual information. We adopt attention-based approach to process memory features where we first learn the spatio-temporal relation among the resultant visual and displacement memory tokens using self-attention module. Further, the output features of self-attention are aggregated with the current visual features through cross-attention. The cross-attended features are finally given to a decoder to predict depth on the current frame. Through extensive experiments on several benchmarks, including KITTI, NYU-Depth V2, and DDAD, we show that MAMo consistently improves monocular depth estimation networks and sets new state-of-the-art (SOTA) accuracy. Notably, our MAMo video depth estimation provides higher accuracy with lower latency, when comparing to SOTA cost-volume-based video depth models.
\end{abstract}

\vspace{-2em}
\section{Introduction}\vspace{-3pt}
Depth plays a fundamental role in 3D perception. Therefore, accurate depth estimation is critical in various applications, such as autonomous driving, AR/VR, and robotics.   
While it is possible to measure depth using LiDAR or Time-of-Flight (ToF) sensors, these sensors are expensive, consume a lot of power, require extensive calibration, and cannot generate reliable measurements for certain surfaces. 
On the other hand, inferring depth from camera images has recently become an cost-efficient and promising alternative. Traditional approaches~\cite{saxena2007depth, furukawa2010towards, newcombe2011dtam} utilize stereo vision and/or structure-from-motion to estimate depth, which, however, have limited accuracy. By leveraging deep learning, researchers have achieved significantly more accurate image-based depth estimation~\cite{eigen2014depth,fu2018deep,bhat2021adabins,ranftl2021vision,yuan2022newcrfs}.

\begin{figure}[!t]
    \centering
    \includegraphics[width=0.98\linewidth]{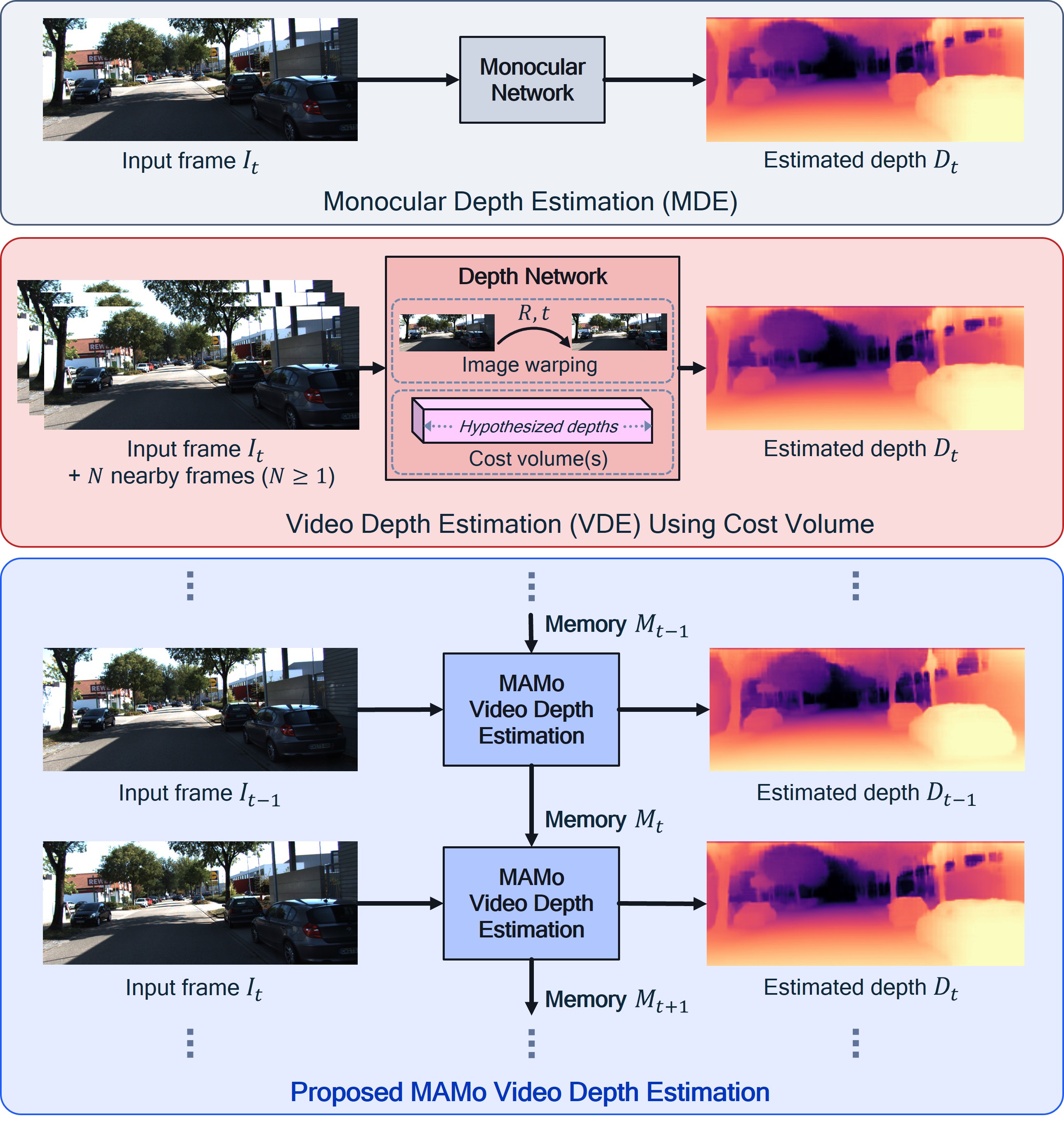}
    \caption{Our proposed MAMo (bottom) enables video depth estimation efficiently in a streaming fashion, by leveraging memory and attention. Monocular depth estimation fails to leverage temporal information (top), while existing cost-volume-based video depth models are computationally expensive (middle). For instance, for each inference, they require multiple image warping operations as well as significant memory usage and heavy computation to construct the cost volume(s).}
    \label{fig:motivation-1}
    \vspace{-15pt}
\end{figure}

Using deep neural networks to infer depth from a single camera image, i.e., monocular depth estimation,\footnote{In this paper, we refer to depth estimation based on a single image as \textit{monocular depth estimation} and depth estimation using consecutive frames captured by the same monocular camera as \textit{video depth estimation}.} has been one of the most popular choices. Monocular depth estimation, however, only predicts depth based on individual images and does not utilize the temporal information from videos, which are almost always available in many applications, e.g., autonomous driving, AR/VR. More recently, researchers have proposed various ways to leverage multiple frames for depth estimation. 
One common approach 
is to utilize a cost volume (or multiple cost volumes), which is used to evaluate depth hypotheses and can be embedded into a deep learning architecture. Cost volumes have enabled 
considerable boost in performance 
at the expense of high computational complexity and memory usage. 
Other works propose video depth estimation models without cost volumes, by leveraging recurrent network~\cite{zhang2019exploiting, patil2020don}, optical flow~\cite{eom2019temporally, xie2020video}, and/or attention~\cite{cao2021learning, wang2022less}. While these models can be more computationally efficient as compared to cost volumes, they have not been shown to provide SOTA accuracy. Moreover, existing video depth estimation methods do not incorporate the latest developments from monocular depth architectures and as a result, they can underperform SOTA monocular depth estimation models despite using more information. 

In this paper, we propose a novel approach, MAMo, for video depth estimation, which leverages memory and attention to make use of the key temporal information contained in a video. MAMo can be combined with any monocular network (e.g., NeWCRFs~\cite{yuan2022newcrfs}, PixelFormer~\cite{Agarwal_2023_WACV}) to perform video depth estimation in a streaming fashion. As such, it is complementary to any existing and future developments in monocular depth estimation. Furthermore, it improves depth estimation accuracy being significantly compute efficient compared to cost volumes. 

Fig.~\ref{fig:motivation-1} (bottom) provides a high-level outline 
of our proposed MAMo framework. We introduce a memory to augment 
the depth estimation process as the network goes through the video frames, which maintains learned visual and displacement tokens storing useful information from a set of consecutive previous frames.
These tokens are cross-referenced using a cross-attention approach when the network derives the depth for the current input frame. 

We propose a novel update scheme for memory module to effectively retain the relevant information from past frames.
More specifically, when performing a memory update, we first predict depths using the current frame and a synthesized version of it warped from the previous frame using optical flow, respectively.
We compare and minimize the difference between the two predictions, and back-propagate the gradients to update the memory, with the depth network's weights frozen. 
Since the memory tokens are used to cross-attend the respective visual features during the two forward passes, they are updated to capture features that is shared across the current frame and the warped previous frame, i.e., the equivariant (\withrespectto~ motion) features across the current and the previous frames.
As we will show, our proposed memory update is more effective compared to 
sliding window style concatenating

Our main contributions are summarized as follows:
\vspace{-0.5em}
\begin{itemize}
    \itemsep0em 
    
    \item We introduce MAMo, a novel memory and attention based framework for video depth estimation.
    MAMo can be combined with any monocular depth network, enabling it to utilize the temporal information to predict more accurate depth.
    
    \item In MAMo, we augment model with memory to retain 
    tokens that capture useful information from the previous frames. These tokens are used to assist depth prediction of the current input frame, via cross-attention. 
    
    \item We propose a novel memory update scheme to effectively retain the relevant information from past frames.
    Specifically, the memory tokens are updated to encode (motion) equivariant features across the current frame and the previous frame.
    
    \item We additionally incorporate careful designs to further improve the video depth estimation performance, such as carrying over decoder features from the previous time step. 
    
    \item We conduct extensive experiments on common depth estimation datasets: KITTI~\cite{kitti}, NYU Depth V2~\cite{silberman2012indoor}, and DDAD~\cite{packnet}. We show that MAMo not only consistently improves latest monocular depth networks, but also outperforms existing SOTA video depth estimation methods. It is also significantly more efficient as compared to approaches that use cost volumes.   
\end{itemize}

\begin{figure*}[t!]
    \centering
    \includegraphics[width=0.8\linewidth]{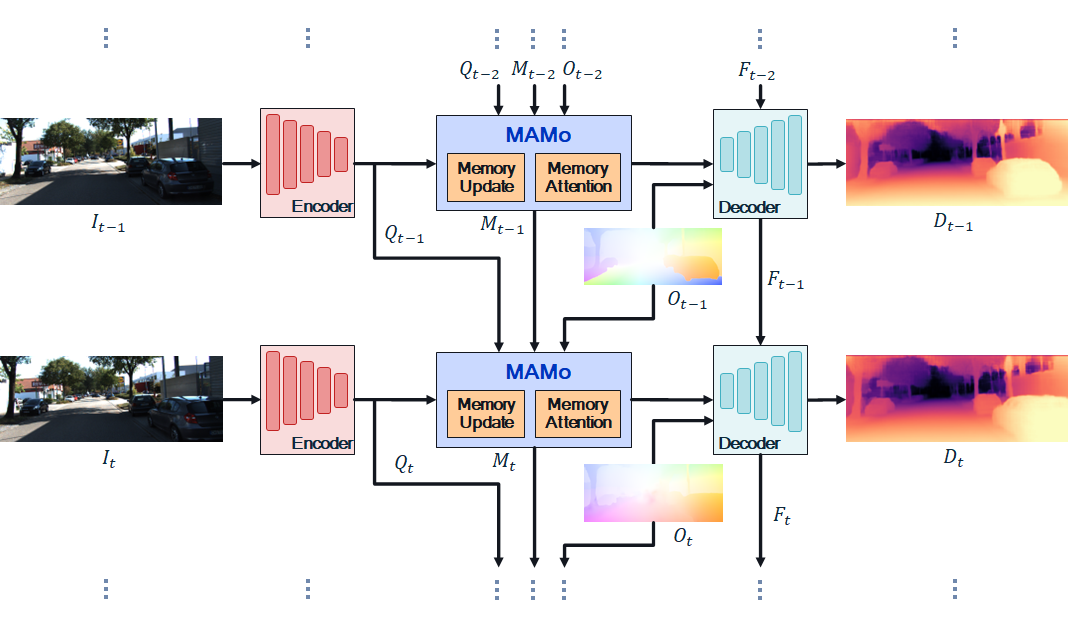} 
    \vskip -5pt 
    \caption{Overview of proposed MAMo method.}
    \label{fig:overview}
    \vspace{-1.0em}
\end{figure*}

\vspace{-10pt}
\section{Related Work}\vspace{-3pt}
\noindent \textbf{Monocular Depth Estimation (MDE):}
 In earlier works, various traditional methods have been proposed for monocular depth estimation~\cite{michels2005high, nagai2002hmm, saxena2005learning, saxena2008make3d, wang2015depth}. 
 Recently, deep-learning-based techniques have gained prominence, which can be broadly categorized into two groups, (i) regressing continuous depth values~\cite{monodepth17,yuan2022newcrfs,zhu1232020mda,cai2021x,shi2023ega} and (ii) treating depth prediction as classification or ordinal regression~\cite{bhat2021adabins,li2022binsformer,fu2018deep}. While researchers continue to investigate monocular depth estimation and improve the accuracy, these methods are fundamentally limited as they cannot leverage temporal information when video data is available.

\vspace{1pt}
\noindent\textbf{Video Depth Estimation (VDE):}
Some existing methods devise networks that predict depth based on more than one frame of the video. 
For instance, ManyDepth~\cite{watson2021temporal} utilizes two consecutive frames by leveraging a cost volume. It is also possible to use more frames via cost volume to perform depth estimation, e.g.,~\cite{Long_2021_CVPR, sayed2022simplerecon}. However, using more frames can result in delays for the depth prediction, and the cost volume architecture is expensive in terms of computational complexity and memory usage. 
Other works explore the use of recurrent neural networks, but only obtain sub-optimal accuracy~\cite{eom2019temporally, zhang2019exploiting, patil2020don}. 
More recently, researchers have started to look into leveraging attention mechanisms for video depth estimation, but the existing methods do not achieve state-of-the-art performance, even when compared to the latest MDE models~\cite{cao2021learning, wang2022less}.
In the context of video depth estimation, it can be useful to utilize optical flow\cite{teed2020raft,huang2022flowformer,sun2018pwc,Garrepalli_2023_CVPR, jeong2022imposing,Jeong_2023_CVPR} information to capture the motion across frames. This has been explored by earlier works~\cite{eom2019temporally,xie2020video}. Additionally, optical flow driven Depth Estimation with use of memory attention is beneficial for robustness, by using motion cues to detect novel objects i.e., openset and OoD \cite{ruff2021unifying} complementing appearance based features, representational learning \cite{pmlr-v80-liu18e,garrepalli2022oracle,liu2022pac,das2023transadapt,Borse_2022_CVPR,Borse_2023_CVPR,vs2023mask,vs2023towards}.

\vspace{1pt}
\noindent \textbf{Memory:}  Use of memory techniques is an extensively researched topic in the NLP community~\cite{graves2014neural,sukhbaatar2015end,weston2014memory}, addressing reasoning tasks like dialogue communication~\cite{wu2019global}, question-answering~\cite{kumar2016ask}, and story generation~\cite{rahman2022make}. Memory is recently introduced in computer vision tasks like image captioning~\cite{chunseong2017attend}, colorization~\cite{yoo2019coloring}, text-to-image synthesis~\cite{zhu2019dm}, video object segmentation~\cite{miao2020memory}, and object tracking~\cite{cai2022memot}. 
Inspired from these methods, we propose MAMo which constructs a memory to maintain relevant spatiotemporal information that can be used to guide depth prediction.

\vspace{-1.em}
\section{Proposed Approach: MAMo}\label{sec:method}
\vspace{-3pt}
In this section, 
we present MAMo, a memory and attention based framework for video depth estimation (VDE).
We provide an overview of MAMo in Section~\ref{sec:mamo} In Sections~\ref{sec:memory_update}~and~\ref{sec:memory_attention}, we provide detailed descriptions of the key components of MAMo, i.e., Memory Update (MU) and Memory Attention (MA). In Section~\ref{sec:additional}, we discuss additional designs that can further enhance video depth estimation. We describe training details in Section~\ref{sec:training}.

\subsection{Using Memory and Attention for VDE}\label{sec:mamo}
\vspace{-3pt}
Consider a sequence of video frames $\{I_0,...,I_t,...,I_{T}\}$. We denote the predicted depths on these frames as $\{D_0,...,D_t,...,D_{T}\}$ and the estimated optical flows between consecutive frame pairs as $\{{O}_1,...,{O}_t,...,{O}_{T}\}$, where $O_t$ is the forward flow from $I_{t-1}$ to $I_t$.\footnote{The optical flows can be estimated by models such as RAFT~\cite{teed2020raft}.} Given an encoder-decoder architecture for depth estimation, we denote the features extracted by the encoder at time $t$ as $Q_t$. We denote the encoder as $h$ and the full depth estimation model as $g$.

Our goal is to develop a depth estimation model that can leverage the temporal correlation across frames as it streams through the video. 
With this motivation, we augment model with memory to retain
a set of learned informative tokens derived from previous $L$ frames as well as the optical flows of the previous $L$ time steps. Formally, $M_t = \{{M}^V_{t}, M^D_t\}$ is the memory at time $t$, and $\{M^V_t\} = \{V_{t-L+1},...,V_{t}\}$ and  $M^D_{t} = \{{P}_{t-L+1},...,{P}_{t}\}$, where $M^V_t$ stores visual information tokens and $M^D_t$ stores pairwise relative displacement tokens based on the previous $L$ time steps. 

At every time step $t$, given the current input frame $I_t$, the optical flow ${O}_{t}$, and the previous memory $M_{t-1}$, MAMo first updates the memory to $M_t$, in order to capture equivariant information across the current and previous time steps. Next, the updated memory tokens goes through self-attention. The processed features are then fused with the encoder features, $Q_t$, through cross-attention. The final aggregated features are fed to the decoder to derive the estimated depth, $D_t$, for the current input frame. 
Additionally, the decoder features from the previous time step, $F_{t-1}$, are carried over to aide the current prediction. We mathematically define the depth prediction as follows:
\begin{equation}
 D_t = g(I_t;\, M_t, O_t, F_{t-1}).
\end{equation}

Fig.~\ref{fig:overview} summarizes our proposed MAMo framework. Overall, it leverages memory and attention mechanisms for video depth estimation. It can be seen that MAMo can readily be implemented on top of any existing monocular depth estimation networks with encoder-decoder architectures. As such, MAMo allows one to take advantage of the latest state-of-the-art and future monocular networks, and convert them to video depth models.

\begin{figure}[t!]
    \centering
    \includegraphics[width=0.8\linewidth]{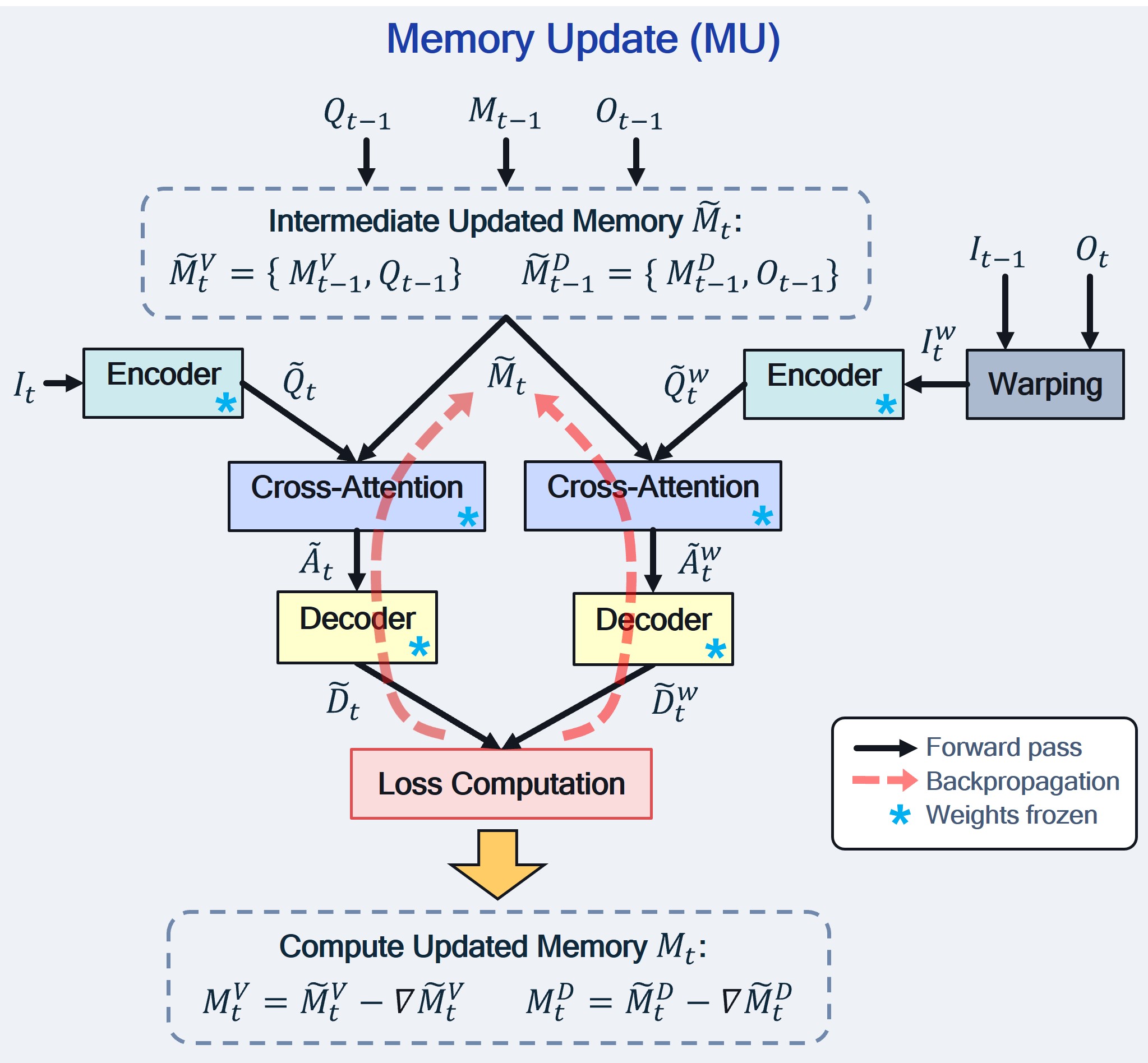} 
    \caption{Overview of proposed memory update scheme. To concisely illustrate the main idea of memory update, we omit some operations in the figure, e.g., self-attention on memory tokens (c.f.~Section~\ref{sec:memory_attention}), decoder feature carry-over (c.f.~Section~\ref{sec:additional}).}
    \label{fig:memory}
    \vspace{-0em}
\end{figure}

\vspace{-7pt}
\subsection{Memory Update}\label{sec:memory_update}
\vspace{-3pt}
As the depth estimation model goes through the video frames, it is critical to appropriately 
update the memory, in order to maintain information that is useful for the current-time depth prediction. As such, we propose a novel scheme to update the memory tokens to capture features that are shared across the current time and the previous time, \thatis~, equivariant features \withrespectto~ motion across the two frames.

More specifically, at time $t$, given the previous encoder features $Q_{t-1}$, memory $M_{t-1}$, and optical flow $O_{t-1}$, we first perform an intermediate update to the memory tokens, \thatis~, $\widetilde{M}_t = \{\widetilde{M}_t^V, \widetilde{M}_t^D\}$, where $\widetilde{M}_t^V = \{M_{t-1}^V, Q_{t-1}\}$ which is the concatenation of previous visual tokens and previous encoder features, and $\widetilde{M}_t^D = \{M_{t-1}^D, O_{t-1}\}$, which is the concatenation of the previous displacement tokens and previous optical flow. After the concatenation, we discard the first tokens in $\widetilde{M}_t^V$ and $\widetilde{M}_t^D$, respectively, so as to maintain the same memory length $L$. 
\setlength{\textfloatsep}{10pt} 

\begin{algorithm}[t]
\caption{Video depth prediction using MAMo}
\label{alg:test}
\footnotesize{
   
\begin{algorithmic}[]
\STATE \textbf{Input}: Video frames $\{I_0,...,I_T\}$ 
\STATE \textbf{Model}: $h(\cdot)$ and $g(\cdot)$: encoder and full depth network
\STATE \textit{\textbf{Initialization}}
\STATE \quad $Q_0 \gets h(I_0),\ \ O_0 \gets \mathbf{0},\ \ F_{-1} \gets \mathbf{0}$ \\
\STATE \quad \textit{Update} $M_{0}$ \quad (repeat $Q_0 \ \ and \ O_0$ for \textit{L} times)
\STATE \quad $D_0 \gets g(I_{0}; M_0, O_0, F_{-1})$

\FOR {$I_t \in \{I_1,...,I_T\}$}
\STATE \quad \textit{Estimate} $O_t$ 
\STATE \textit{\textbf{Memory Update}} \ \ (Sec.~\ref{sec:memory_update})
\STATE \quad $\widetilde{M}_t^V \gets \{M_{t-1}^V, Q_{t-1}\}, \ \ \widetilde{M}_t^D \gets \{M_{t-1}^D, O_{t-1}\}$ \\
\STATE \quad $\widetilde{M}_t \gets \{\widetilde{M}_t^V, \widetilde{M}_t^D\}$\\
\STATE \quad $I_t^w \gets \textit{Warp}(I_{t-1}, O_t)$\\
\STATE \quad $\widetilde{D}_{t} \gets g(I_t;\widetilde{M}_t, O_t, F_{t-1})$\\
\STATE \quad  $\widetilde{D}_t^w \gets g(I_t^w;\widetilde{M}_t, O_t, F_{t-1})$\\

\STATE \quad \textit{SILogLoss} ($\widetilde{D}_{t}$, $\widetilde{D}_t^w$)\\
\STATE \quad \textit{Backpropagation}\\
    
            
\STATE \quad \textit{Update} $M_t$ \ \ (Eq.~\ref{eq:mt})\\

\STATE \textit{\textbf{Depth Estimation}}\\
\STATE \quad $D_t \gets g(I_t; M_t, O_t, F_{t-1})$,\quad $Q_t \gets h(I_t)$
\ENDFOR
\end{algorithmic}
}
\end{algorithm}

Next, we perform two forward passes of the depth network, with the network parameters frozen and both using the intermediate updated memory. In the first pass, we use the current frame $I_t$ as input and the network predicts a depth map $\widetilde{D}_t$. In the second pass, we construct an input frame, $I_t^w$, which is warped from $I_{t-1}$ using $O_t$. In other words, $I_t^w$ is the synthesized version of $I_t$ with motion compensation from $O_t$.
The network consumes $I_t^w$ and generates a depth map $\widetilde{D}_t^w$. We then compute a loss between $\widetilde{D}_t$ and $\widetilde{D}_t^w$ using the Scale-Invariant Logarithmic (SILog) Loss~\cite{eigen2014depth}, and backpropagate the gradients to update the memory tokens:
\begin{equation}
\label{eq:mt}
    {M}_{t}^V = \widetilde{M}_{t}^V - \nabla\widetilde{M}_{t}^V,\quad
    {M}_{t}^D = \widetilde{M}_{t}^D - \nabla\widetilde{M}_{i}^D,
\end{equation}
where $M_t = \{{M}_{t}^V, {M}_{t}^D\}$ is the updated memory that can then be used to predict the depth map $D_t$ for $I_t$.

Fig.~\ref{fig:memory} provides a visual illustration of the memory update steps. During the two forward passes, the memory tokens cross-attend the encoder features $\widetilde{Q}_t$ extracted from $I_t$ and the encoder features $\widetilde{Q}_t^w$ extracted from $I_t^w$, respectively. The resulting cross-attended features, $\widetilde{A}_t$ and $\widetilde{A}_t^w$, are then fed into the decoder to generate depth predictions, $\widetilde{D}_t$ and $\widetilde{D}_t^w$. 
When we minimize the difference between two outputs, the memory is encouraged to capture features that are shared across $Q_t$ and $Q_t^w$. This would make $\widetilde{A}_t$ and $\widetilde{A}_t^w$ more similar since the memory modulates the encoder features via cross-attention, and as a result, $\widetilde{D}_t$ and $\widetilde{D}_t^w$ would become more similar.

By performing the update, the memory tokens learn to keep similar features shared by $I_t$ and $I_t^w$. In other words, these are the equivariant features across $I_t$ and $I_{t-1}$ \withrespectto~ the optical flow. Additionally, this update mechanism potentially suppresses the noisy inconsistencies in memory. 
Overall, our memory and attention framework allows MAMo to be temporally consistent \withrespectto~ motion equivariant features, while also implicitly learning to better filter and aggregate spatio-temporal information for non-equivariant/inconsistent regions towards smoother and consistent VDE 
and hence performing better than sliding window style motion compensated concatenation. 
We summarize the inference procedure of our proposed MAMo video depth estimation in Algorithm~\ref{alg:test}.

\begin{table*}[t!]
    \caption{Quantitative results on KITTI (Eigen split) for distances up to 80 meters. $\dagger$ means methods uses multiple networks to estimate depth. ManyDepth-FS, and TC-Depth-FS means ManyDepth and TC-Depth are trained in fully-supervised fashion using ground-truths respectively. MF means multi frame methods, SF means single frame methods, and VD means extending MDE to VDE methods.$\uparrow$ means higher the better, and $\downarrow$ means lower the better. }
	\label{tab:KITTI}
    \adjustbox{max width=1\textwidth}
    {
    \begin{tabular}{ll|l|ccccccc}
    \hline
        Type & Method & Encoder & Abs Rel$\downarrow$  & Sq Rel$\downarrow$ & RMSE$\downarrow$ & $\text{RMSE}_{log}\downarrow$ & $\delta<1.25\uparrow$ & $\delta<1.25^2\uparrow$ & $\delta<1.25^3\uparrow$  \\ \hline
        \multirow{12}{*}{MF} & NeuralRGB~\cite{liu2019neural} &  CNN based$\dagger$ & 0.100 & -- & 2.829 & --  & 0.931 & -- & --\\ 
        & ST-CLSTM~\cite{zhang2019exploiting} & Resnet18 & 0.101 & -- & 4.137 & --  & 0.890 & 0.970 & 0.9890\\ 
        & FlowGRU~\cite{eom2019temporally} & CNN~\cite{eom2019temporally} & 0.112 & 0.700 & 4.260 & 0.184  & 0.881 & 0.962 & 0.9830\\ 
        & Flow2Depth~\cite{xie2020video} &  CNN~\cite{mayer2016large}$\dagger$ & 0.081 & 0.488 & 3.651 & 0.146  & 0.912 & 0.970 & 0.9883\\ 
        & RDE-MV~\cite{patil2020don} &  ResNet18$\dagger$ & 0.111 & 0.821 & 4.650 & 0.187  & 0.821 & 0.961 & 0.9823\\ 
        & Patil \textit{et.al.}~\cite{patil2020don} &  ResNet18$\dagger$+ConvLSTM & 0.102 & -- & 4.148 & --  & 0.884 & 0.961 & 0.9824\\
        & Cao \textit{et.al.}~\cite{cao2021learning} &  -- & 0.099 & -- & 3.832 & --  & 0.886 & 0.968 & 0.9890\\ 
        & STAD~\cite{lee2021stad} & CNN $\dagger$~\cite{liu2019neural} & 0.109 & 0.594 & 3.312 & 0.153  & 0.889 & 0.971 & 0.9890\\ 
        & FMNet~\cite{wang2022less} &  ResNeXt-101 & 0.099 & -- & 3.832 & 0.129  & 0.886 & 0.968 & 0.9893\\  
        & ManyDepth-FS~\cite{watson2021temporal} &  ResNet50 & 0.069 & 0.342 & 3.414 & 0.111  & 0.930 & 0.989 & 0.9970\\
        & ManyDepth-FS~\cite{watson2021temporal} &  Swin-large & 0.060 & 0.248 & 2.747 & 0.099  & 0.955 & 0.993 & 0.9981\\
        & TC-Depth-FS~\cite{ruhkamp2021attention} &  ResNet50 & 0.071 & 0.330 & 3.222 & 0.108  & 0.922 & 0.993 & 0.9970\\ \hline
        \multirow{3}{*}{SF} & AdaBins~\cite{bhat2021adabins} &  EfficientNet-B5+mViT~\cite{tan2019efficientnet} &0.058	& 0.190	& 2.360	&0.088	&0.964	&0.995	&0.9991\\
        & BinsFormer~\cite{li2022binsformer} & Swin-large &0.052	&0.151	&2.098	&0.079&0.975	&0.997	&0.9992	\\	
        & DepthFormer~\cite{agarwal2022depthformer} & MiT-B4~\cite{xie2021segformer} &0.058	&0.187	& 2.285	&0.087	&0.967	&0.996	&0.9991 \\ 
        \hline
        \multirow{8}{*}{VD} &  ResNet-DPT &  ResNet50 & 0.085 & 0.383 & 3.242 & 0.130 & 0.913 & 0.981  & 0.9960 \\
        & \cellcolor{lightgray}ResNet-DPT+MAMo (ours) &  \cellcolor{lightgray}ResNet50 & \cellcolor{lightgray}0.071 & \cellcolor{lightgray}0.301 & \cellcolor{lightgray}2.984 & \cellcolor{lightgray}0.121  & \cellcolor{lightgray}0.926 & \cellcolor{lightgray}0.990&
        \cellcolor{lightgray}0.9971 \\
        & NeWCRFs~\cite{yuan2022newcrfs} & Swin-Base & 0.054 & 0.157 & 2.140 & 0.081 & 0.973 & 0.997 & 0.9993  \\ 
        & \cellcolor{lightgray} NeWCRFs+MAMo (ours)  & \cellcolor{lightgray} Swin-Base & \cellcolor{lightgray} 0.051 & \cellcolor{lightgray} 0.149 & \cellcolor{lightgray} 2.090 & \cellcolor{lightgray} 0.078 & \cellcolor{lightgray} 0.976 & \cellcolor{lightgray} 0.998 & \cellcolor{lightgray} 0.9994  \\ 
        & NeWCRFs & Swin-large & 0.053 & 0.154 & 2.118 & 0.080 & 0.974 & 0.997 & 0.9994  \\
        & \cellcolor{lightgray} NeWCRFs+MAMo (ours) & \cellcolor{lightgray} Swin-large & \cellcolor{lightgray} 0.050 & \cellcolor{lightgray} 0.141 & \cellcolor{lightgray} 2.003 & \cellcolor{lightgray} 0.076 & \cellcolor{lightgray} \textbf{0.977} & \cellcolor{lightgray} \textbf{0.998} & \cellcolor{lightgray} 0.9994  \\ 
        & PixelFormer~\cite{Agarwal_2023_WACV} & Swin-large & 0.052 & 0.152 & 2.093 & 0.079 & 0.975 & 0.997 & 0.9994  \\
        & \cellcolor{lightgray} PixelFormer+MAMo (ours) & \cellcolor{lightgray} Swin-large & \cellcolor{lightgray} \textbf{0.049} & \cellcolor{lightgray} \textbf{0.130} & \cellcolor{lightgray} \textbf{1.884} & \cellcolor{lightgray} \textbf{0.072} & \cellcolor{lightgray} \textbf{0.977} & \cellcolor{lightgray} \textbf{0.998} & \cellcolor{lightgray} \textbf{0.9995}  \\
        \hline   
    \end{tabular}
    }
    \vspace{-1em}
\end{table*}

\vspace{-0.8em}
\subsection{Memory Attention}\label{sec:memory_attention}
\vspace{-4pt}
We adopt attention-mechanisms to process updated memory features $M_t$
for the actual depth prediction at time $t$. 
First, we perform self-attention over the visual memory tokens $M_t^V$ and also derive corresponding positional encodings from the displacement memory tokens $M_t^D$. More specifically, we feed $M_t^P$, which contains the past pairwise optical flow information, into a convolutional block. 
Linear weighting within convolutional layer operations learns to approximate aggregate estimate of
relative motion between the current time and each previous time step tracked in the memory. By doing this, we do not need to explicitly calculate the optical flow between the current time and each of previous time steps 
which would be computationally demanding. More formally,
\begin{equation}\vspace{-1pt}
    A_t^\text{self} = \text{SelfAttn}(M_t^V;\, \text{Conv}(M_t^D)),
\end{equation}
where $A_t^\text{self}$ are the output features from the self-attention module, $\text{SelfAttn}(x;\, y)$ denotes self-attention over $x$ with positional encodings of $y$, and $\text{Conv}(.)$ denotes convolutional layers.
Next, the self-attended memory features modulate the encoder features of the current frame via cross-attention:\vspace{-1pt}
\begin{equation}
    A_t = \text{CrossAttn}(A_t^\text{self},\, Q_t),
\end{equation}
where $\text{CrossAttn}(.,.)$ denotes the cross-attention operation and $A_t$ then goes into the decoder for final depth prediction.


\vspace{-1pt}
\subsection{Additional Improvements}\label{sec:additional}
\vspace{-3pt}
To further enable the depth network to utilize available temporal information, we carry over the previous decoder features $F_{t-1}$ and provide them as part of the input to the decoder at time $t$. In additional the optical flow, $O_t$, between the previous frame and the current frame is also supplied to the decoder. In this way, the decoder is aware of the relative pixel-wise motion from $t-1$ to $t$ and can thus learn to properly incorporate the previous features for the current depth prediction. While our proposed MAMo approach does not require these additional designs to work well, they do provide further improvements on depth prediction accuracy, as we will show in the experiments.

\vspace{-2pt}
\subsection{Training} \label{sec:training}
\vspace{-4pt}
In training, at every time-step, we first perform the inference to compute depth (c.f.~Algorithm~\ref{alg:test}), which is then compared with ground-truth depth using SILog loss~\cite{eigen2014depth}:
\begin{equation}
    \mathcal{L}_d = \alpha\sqrt{\frac{1}{n}\sum_k\left(\delta d_k\right)^2-\frac{\lambda}{n^2}\left(\sum_k\delta d_k\right)^2},
\end{equation}
where $\delta d_k = \log{{D}_{t}}(k)-\log{{D}^{gt}_{t}(k)}$, $D_{t}^{gt}$ is the ground-truth depth for $I_t$, $k$ is the pixel location, $n$ is the total number of pixels, and $\alpha = 10$ and $\lambda = 0.85$ following~\cite{eigen2014depth}.  

In order to allow the network to train on more motion situations, we employ a video augmentation strategy via subsampling. More specifically, we use subsampled sequences of length $T$ for training, i.e., $\{I_0,I_r,...,I_{t\times r},...,I_{T\times r}\}$, where $r$ is a sub-sampling ratio randomly selected between 1 and 4 at every epoch for each sequence. As an example, if $r = 4$ and $T=8$, the video sequence is $\{I_0,I_4,I_{8},...,I_{32}\}$, this allows the network to see larger motion across the frames. Effectively, this augmentation increases the maximum video frame range to $4\times T$.


\vspace{-1pt}
\section{Experiments}\vspace{-4pt}
\subsection{Implementation}\vspace{-3pt}
\noindent \textbf{Networks:} We use MAMo to enable latest SOTA monocular methods, e.g., PixelFormer~\cite{Agarwal_2023_WACV}, NeWCRFs~\cite{yuan2022newcrfs}, as well as a strong convolutional baseline, i.e., a variant of DPT~\cite{ranftl2021vision} with a ResNet encoder (referred to as ResNet-DPT), to perform video depth estimation. Since all these monocular models have encoder-decoder architectures, MAMo can be readily applied. For PixelFormer and NeWCRFs, we extend their own attention designs to create the memory attention modules, respectively. We use Linformer~\cite{wang2020linformer} to create the self- and cross-attention blocks for ResNet-DPT.\footnote{See supplementary file for more details on how we apply MAMo to these models.} 
To obtain optical flow estimation, we use RAFT~\cite{teed2020raft} for our main results. We also conduct ablation study using the lightweight RAFT-Small model.


\vspace{1pt}
\noindent \textbf{Hyperparameters:} In all our experiments, we set $T = 8$ and $L = 4$ unless otherwise mentioned. Given the input frame $I_t$ of size $H\times W$, we set the visual memory token size to $512 \times \frac{H}{32} \times \frac{W}{32}$. Thus, the size of $M_t^V$ is $L \times 512 \times \frac{H}{32} \times \frac{W}{32} $ and the size of $M_t^D$ is $L \times 2 \times H \times W $. 

\vspace{1pt}
\noindent \textbf{Training:} We perform all our experiments using 4 NVIDIA-V100 GPUs. We train the network for 25 epochs, using Adam optimizer with a batch size of 8. We set the initial learning rate to $4\times 10^5$ and then linearly decrease it to $4\times 10^6$ across the training iterations.

\vspace{1pt}
\noindent \textbf{Evaluation:} We use the standard metrics to evaluate depth estimation results; see~\cite{eigen2014depth} for metric definitions.

\subsection{Datasets}
\vspace{-3pt}
\textbf{KITTI~\cite{kitti}:} KITTI is one of the most commonly used benchmarks for outdoor depth estimation. We use the Eigen split~\cite{eigen2014depth} for training and testing, which has 23,488 training images and 697 test images. When training and testing our proposed MAMo approach as well as existing video depth models (e.g., ManyDepth~\cite{watson2021temporal}), we use the video (sub)sequences that correspond to the training and test images. The video frames are 375$\!\times\!$1241 and the depth range is 80 meters.

\textbf{DDAD~\cite{packnet}:} Dense Depth for Autonomous Driving (DDAD) is very recent dataset featuring urban driving scenarios and long ranges (up to 250 meters). It contains 12,650 training and 3,950 validation samples. We conduct zero-shot transfer to evaluate the generalizability of the trained models from KITTI on all 3,950 validation samples.

\textbf{NYU Depth V2~\cite{silberman2012indoor}:} 
This is a standard dataset for indoor depth estimation, containing 120K RGB-D videos captured from 464 indoor scenes. Since the original test set only contains individual images, we create training and test splits for the video setting. Specifically, from the original 249 training scenes proposed in~\cite{eigen2014depth}, we use 198 scenes (25,342 image-depth pairs) for training and 86 scenes (10,911 test images) for testing. We refer this video depth version as \textit{NYUDv2-Video}. The images are of size 480$\!\times\!$640 with a maximum depth range of 10 meters.

\begin{table}[t!]
    \caption{Quantitative results on DDAD dataset for distances up to 200 meters, and input frame resolution is $1216\times 1936$.}
	\label{tab:DDAD}
    \centering
    \resizebox{1\linewidth}{!}{
    \small
    \begin{tabular}{l|l|ccc}
    \hline
        Method & Encoder & Sq Rel$\downarrow$ & RMSE$\downarrow$ & $\delta<1.25\uparrow$ \\ \hline
        ManyDepth-FS~\cite{watson2021temporal} & ResNet50 & 5.471& 16.123  & 0.744 \\ 
        ManyDepth-FS~\cite{watson2021temporal} & Swin-large & 4.211 & 13.899 & 0.784 \\ 
        TC-Depth-FS~\cite{ruhkamp2021attention} & ResNet50 & 5.285 & 15.121 & 0.777 \\ 
        AdaBins~\cite{bhat2021adabins} & \cite{tan2019efficientnet}  & 4.950 &  15.228	&  0.780 \\
        DepthFormer~\cite{agarwal2022depthformer} & MiT-B4~\cite{xie2021segformer} & 4.791	& 14.595	& 0.789	\\
        \hline
        ResNet-DPT & ResNet50 & 5.874 & 15.697 &  0.761 \\ 
        \rowcolor{lightgray} ResNet-DPT+MAMo (ours) & ResNet50 & 4.659 & 14.288 &  0.804\\ 
        NeWCRFs & Swin-large & 4.041 & 11.956 & 0.816 \\ 
        \rowcolor{lightgray} NeWCRFs+MAMo (ours) & Swin-large & \textbf{2.990} & \textbf{10.462} & 0.867 \\  
        PixelFormer &  Swin-large & 4.474& 12.467 & 0.802 \\ 
        \rowcolor{lightgray} PixelFormer+MAMo (ours) & Swin-large & 3.349 & 11.094 & \textbf{0.870}    \\ 
        \hline
    \end{tabular}
    }
    \vspace{-0.0em}
\end{table}

\begin{table*}[t!]
    \caption{Quantitative results on NYUv2-Video dataset. 
    }
	\label{tab:NYUv2-sec4.4}
    \centering
    \resizebox{1\linewidth}{!}{
    \small
    \begin{tabular}{l|l|ccccccc}
    \hline
        Method & Encoder & Abs Rel$\downarrow$  & Sq Rel$\downarrow$ & RMSE$\downarrow$ & $\text{RMSE}_{log}\downarrow$  & $\delta<1.25\uparrow$ & $\delta<1.25^2\uparrow$ & $\delta<1.25^3\uparrow$ \\ \hline
        ResNet-DPT &  ResNet50 & 0.127 & 0.083 & 0.464  & 0.157 & 0.846 & 0.974  & 0.9946 \\ 
        \rowcolor{lightgray} ResNet-DPT + MAMo (ours) &  ResNet50 & 0.119 & 0.075 & 0.420 & 0.135 & 0.861 & 0.982  & 0.9963  \\ 
        NeWCRFs & Swin-Base & 0.103 & 0.056 & 0.384 & 0.129  & 0.900 & 0.987 & 0.9978\\ 
        \rowcolor{lightgray} NeWCRFs + MAMo (ours) & Swin-Base&  0.100 & 0.052 & 0.371 & 0.126 & 0.906 & 0.989 & 0.9983 \\ 
        NeWCRFs & Swin-large & 0.097 & 0.051 & 0.365 & 0.123 & 0.912 & 0.989 & 0.9981\\ 
        \rowcolor{lightgray} NeWCRFs + MAMo (ours) & Swin-large & 0.094 & 0.048 & 0.353 & 0.120 & 0.918 & 0.991 & 0.9983\\ 
        PixelFormer &  Swin-large & 0.098 & 0.053 & 0.369 & 0.124 & 0.911 & 0.988 & 0.9983\\ 
        \rowcolor{lightgray} PixelFormer + MAMo (ours) &  Swin-large & \textbf{0.094} & \textbf{0.049} &  \textbf{0.358} & \textbf{0.119} & \textbf{0.919} & \textbf{0.991}  & \textbf{0.9984} \\ 
        \hline
    \end{tabular}
    }
    \vspace{-0em}
\end{table*}

\begin{table*}[t!]
    \caption{Ablation comparisons on NYUv2-Video dataset. Here, MA refers to memory attention, MU refers to memory update, $OF_d$ indicates that optical flow is used as input to decoder, and $F_{t-1}$ denotes previous decoder features. $OF_m$ indicates that optical is used to construct memory $M_t$. Here we use Swin-Base as the encoder for all variants. 
    }
	\label{tab:Abl_NYUv2}
    \centering
    \resizebox{1\linewidth}{!}{
    \small
    \begin{tabular}{l|lllll|ccccc}
    \hline
        Method & $F_{i-1}$ &  $OF_d$ &  $OF_m$ & MA & MU & Abs Rel$\downarrow$  & Sq Rel$\downarrow$ & RMSE$\downarrow$   & $\delta<1.25\uparrow$ & $\delta<1.25^2\uparrow$  \\ \hline
        
        NeWCRFs & & & & & & 0.1029 & 0.0559 & 0.3838   & 0.9004 & 0.9872 \\
        NeWCRFs + $F_{t-1}$& $\checkmark$& & & & & 0.1014 & 0.0539 & 0.3782 & 0.9021 & 0.9882  \\ 
        NeWCRFs + warped $F_{i-1}$ using $O_t$  & $\checkmark$& $\checkmark$& & &  & 0.1018 & 0.0551 & 0.3779  & 0.9032 & 0.9875 \\ 
        NeWCRFs + $O_t$ and $F_{t-1}$ to decoder & $\checkmark$& $\checkmark$  & & & &  0.1009 & 0.0535 & 0.3743 & 0.9049 & 0.9878 \\
        NeWCRFs + sliding window & & & $\checkmark$ & $\checkmark$ & & 0.1024 & 0.0555 & 0.3827 &  0.9002 & 0.9866 \\ 
        NeWCRFs + MU & &  & $\checkmark$ & $\checkmark$ & $\checkmark$& 0.1004 & 0.0527 & 0.3722 & 0.9051 & 0.9888 \\ 
        NeWCRFs + MAMo (ours) & $\checkmark$&$\checkmark$ & $\checkmark$ & $\checkmark$ & $\checkmark$& \textbf{0.1000} & \textbf{0.0524} & \textbf{0.3708}  & \textbf{0.9053} & \textbf{0.9889}  \\ \hline
        
    \end{tabular}
    }
    \vspace{-1.0em}
\end{table*}

\subsection{Results on KITTI} \vspace{-4pt}
Table~\ref{tab:KITTI} shows the results on KITTI. We see that MAMo considerably improves upon ResNet-DPT, NewCRFs, and PixelFormer consistently. 
We additionally compare our MAMo-based models with existing SOTA monocular depth estimation methods, as well as multi-frame or video depth estimation methods. 
Note that we retrain the SOTA cost-volume-based and attention-based multi-frame models of ManyDepth~\cite{watson2021temporal} and TC-Depth~\cite{ruhkamp2021attention} in the supervised setting for a fair comparison; we refer to the supervised versions as ManyDepth-FS and TC-Depth-FS.\footnote{We use the provided supervised learning settings in the original repos.}
It can be seen that our MAMo-based models achieve the SOTA performance and in particular, using MAMo with PixelFormer achieves the best accuracy on KITTI Eigen test set.\footnote{A more comprehensive comparison table including latest, not officially published methods can be found in the supplementary file.}


\subsection{Results on DDAD} \vspace{-4pt}
In Table~\ref{tab:DDAD}, we test the KITTI-trained models on DDAD to evaluate the generalization performance. 
In can be seen that in this case, MAMo still consistently improves the base monocular depth networks. For instance, the squared relative error reduces significantly from 4.041 to 2.990 when applying MAMo to NeWCRFs.
When comparing to the other SOTA monocular models as well as the SOTA multi-frame models of ManyDepth-FS and TC-Depth-FS, our MAMo models have the best depth estimation accuracy. This shows that  MAMo framework enables the networks to properly utilize the temporal information in the video, allowing them to provide superior generalization ability.


\subsection{Results on NYUv2-Video} \vspace{-4pt}
To assess the benefits of using our proposed MAMo in indoor scenarios, we train and evaluate the performance of PixelFormer~\cite{Agarwal_2023_WACV}, NeWCRFs~\cite{yuan2022newcrfs}, and ResNet-DPT~\cite{ranftl2021vision}, with and without MAMo on the NYUv2-Video dataset. Table~\ref{tab:NYUv2-sec4.4} shows that in the indoor setting, our proposed MAMo approach is useful for improve the accuracy of monocular models by properly leveraging the video information. 



\begin{figure*}[t]
    \centering
    \includegraphics[width=0.95\linewidth]{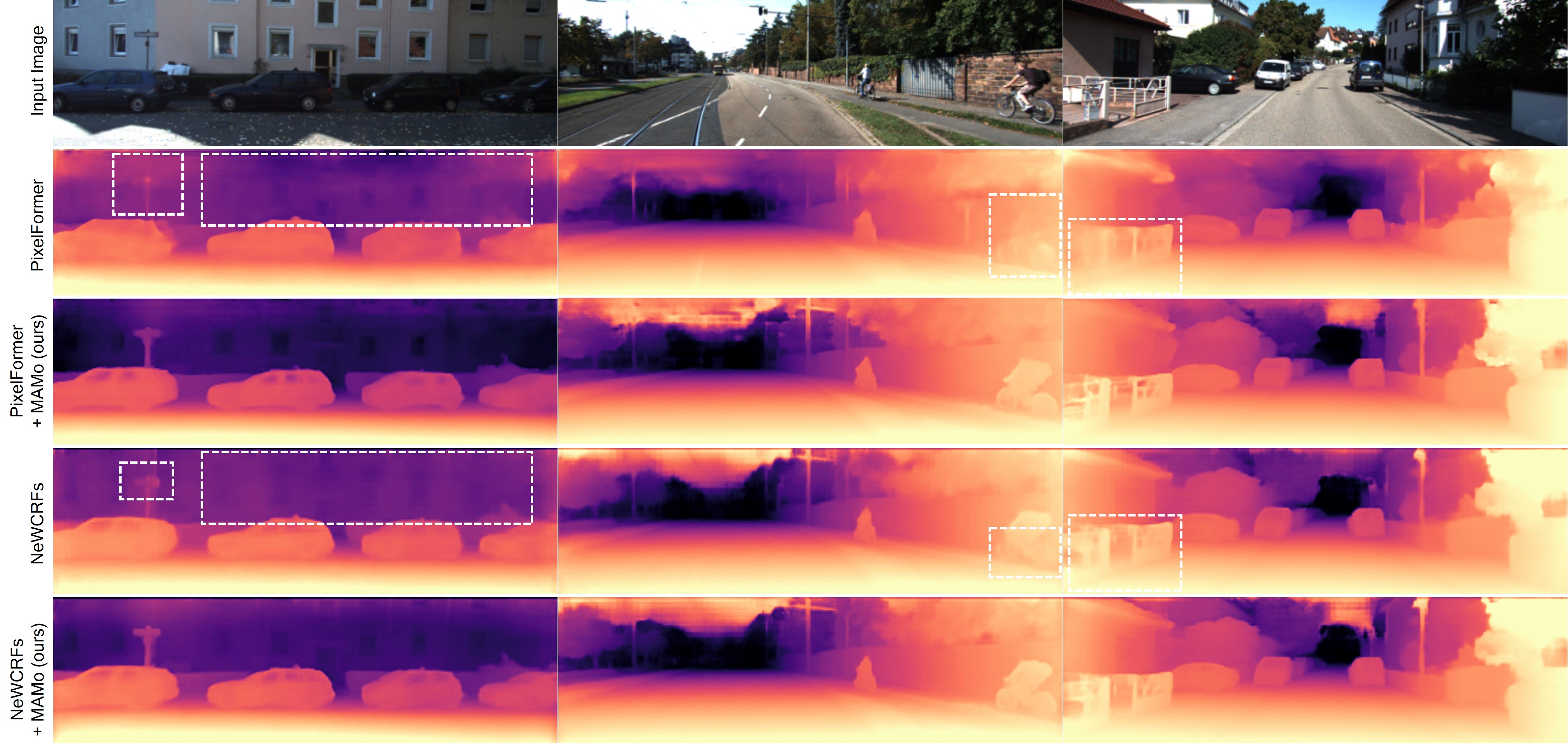}
    \caption{Qualitative results on KITTI. We highlight (white boxes) regions where MAMo significantly improves depth estimation quality.}
    \vspace{-1pt}
    \label{fig:vissualizations}
    \vspace{-0.9em}
\end{figure*}

\begin{table}[t!]
    \caption{Using different optical flow networks for MAMo,on NYUv2-Video dataset. We perform this experiment using NeWCRFs + MAMo with Swin-Large encoder.}
	\label{tab:Abl_OF}
    \centering
    \resizebox{1\linewidth}{!}{
    \normalsize
    \begin{tabular}{l|ccccc}
    \hline
        OF Method & Abs Rel$\downarrow$  & Sq Rel$\downarrow$ & RMSE$\downarrow$  & $\delta<1.25\uparrow$ & $\delta<1.25^2\uparrow$ \\ \hline
         RAFT-small & 0.095  & 0.048& 0.355  & 0.917 &  0.990   \\ 
         RAFT  & 0.094 & 0.048 & 0.353 & 0.918 & 0.991 \\ 
        \hline
    \end{tabular}
    }
    \vspace{-0em}
\end{table}

\section{Discussion}
\vspace{-3pt}
\subsection{Ablation Study} \vspace{-3pt}
We conduct extensive ablation studies to analyze different aspects of our proposed approach and the design choices. 
We conduct experiments on NYUv2-Video using NeWCRFs as the base method, with the Swin-Base encoder. 
As shown in Table~\ref{tab:Abl_NYUv2}, we compare the following options: (1) original NeWCRFs, (2) NeWCRFs + $F_{t-1}$: previous decoder features are used, (3) NeWCRFs + warp $F_{i-1}$ using $O_t$: we warp $F_{t-1}$ using $O_t$ before passing it to the decoder, (4) NeWCRFs + $O_t$ and $F_{t-1}$ to the decoder: $O_t$ and $F_{t-1}$ are concatenated along with encoder features as input to the decoder, (5) NeWCRFs + sliding window: we construct $M_t$ using sliding-window technique to save previous features and optical flows, and then use attention to fuse them with $Q_t$, (6) NeWCRFs + MU: we update $M_t$ using our proposed memory update scheme (c.f. Section~\ref{sec:memory_update}), before feeding it to memory attention and decoder, (7) NeWCRFs + MAMo: applying our full proposed MAMo to NewCRFs.


\begin{table}[t!]
    \caption{Using different memory length $L$ for MAMo. We perform this study using NeWCRFs + MAMo with Swin-Large encoder.}
    \normalsize
	\label{tab:Abl_L}
    \centering
    \resizebox{1\linewidth}{!}{
    \begin{tabular}{l|c|ccccc}
    \hline
        datset & $L$ & Abs Rel$\downarrow$  & Sq Rel$\downarrow$ & RMSE$\downarrow$  & $\delta<1.25\uparrow$ & $\delta<1.25^2\uparrow$ \\ \hline
         \multirow{3}{*}{KITTI} & 2 & 0.051 & 0.149 & 2.030 & 0.976 & 0.998    \\ 
         & 4  & 0.050 & 0.141 & 2.003  & 0.977 & 0.998 \\
         & 6 & \textbf{0.050} & \textbf{0.140} & \textbf{1.990} & \textbf{0.977} & \textbf{0.999} \\ \hline
         \multirow{2}{*}{NYUv2} & 2 & 0.095 & 0.049 & 0.358 & 0.917 & 0.990   \\ 
         \multirow{2}{*}{(video)} & 4  & 0.094 & \textbf{0.048} & 0.353 & 0.918 & \textbf{0.991} \\
          & 6 & \textbf{0.093} & \textbf{0.048} & \textbf{0.351} & \textbf{0.920} & \textbf{0.991}   \\ 
        \hline
    \end{tabular}
    }
\end{table}

It can be seen in Table~\ref{tab:Abl_NYUv2}, carrying over $F_{t-1}$ provides a $1.5\%$ decrease in RMSE. On the other hand, warping $F_{t-1}$ using $O_t$ is not helpful and incurs additional computation to warp the features at multiple scales. Using $F_{t-1}$ and $O_t$ as input to decoder via concatenation, however, enables the decoder to learn efficient positional or motion cues for $F_{t-1}$ \withrespectto~ $I_t$, thus decreasing RMSE by $2.5\%$. Constructing the memory $M_t$ naively with sliding window does not improve the depth estimation accuracy, since this approach keeps irrelevant features in $M_t$, making it more difficult for attention to fuse $M_t$ with $Q_t$; sliding window results in lower accuracy as shown in Table~\ref{tab:Abl_NYUv2}. In contrast, our proposed memory update approach (c.f. Section~\ref{sec:memory_update}) allows the memory to maintain only highly correlated information and reduces RMSE by $3.5\%$ and squared relative error by $6\%$. Finally, combining all our proposed techniques, our full MAMo approach considerably decreases the squared relative error by $6.5\%$ and RMSE by $4\%$ when comparing to the  NeWCRFs.


\vspace{1pt}
\noindent \textbf{Optical Flow:} While we use RAFT to estimate optical flow in our main experiments, MAMo still works well with lighter, more efficient optical flow networks. 
To show this, we compare the depth estimation performance of MAMo by using optical flows generated by RAFT and RAFT-Small. RAFT-Small has significantly lower latency; see \cite{teed2020raft} for detailed comparison.
It can be seen in Table~\ref{tab:Abl_OF}, when using lighter-weight optical flow models like RAFT-small, the depth estimation performance is almost the same as that of using RAFT. 

\noindent \textbf{Memory length $L$:} We perform experiment to ablate on different memory lengths $L$ (e.g., 2, 4, and 6) on KITTI and NYUv2-Video. Note that we set $T=8$ in these experiments. In Table~\ref{tab:Abl_L}, we can see the depth estimation accuracy considerably improves from $L=2$ to $L=4$. Increasing it further to $L=6$ provides additional minor improvements.

\begin{figure}[t!]
    \vspace{-9pt}
    \centering
    \includegraphics[width=0.7\linewidth]{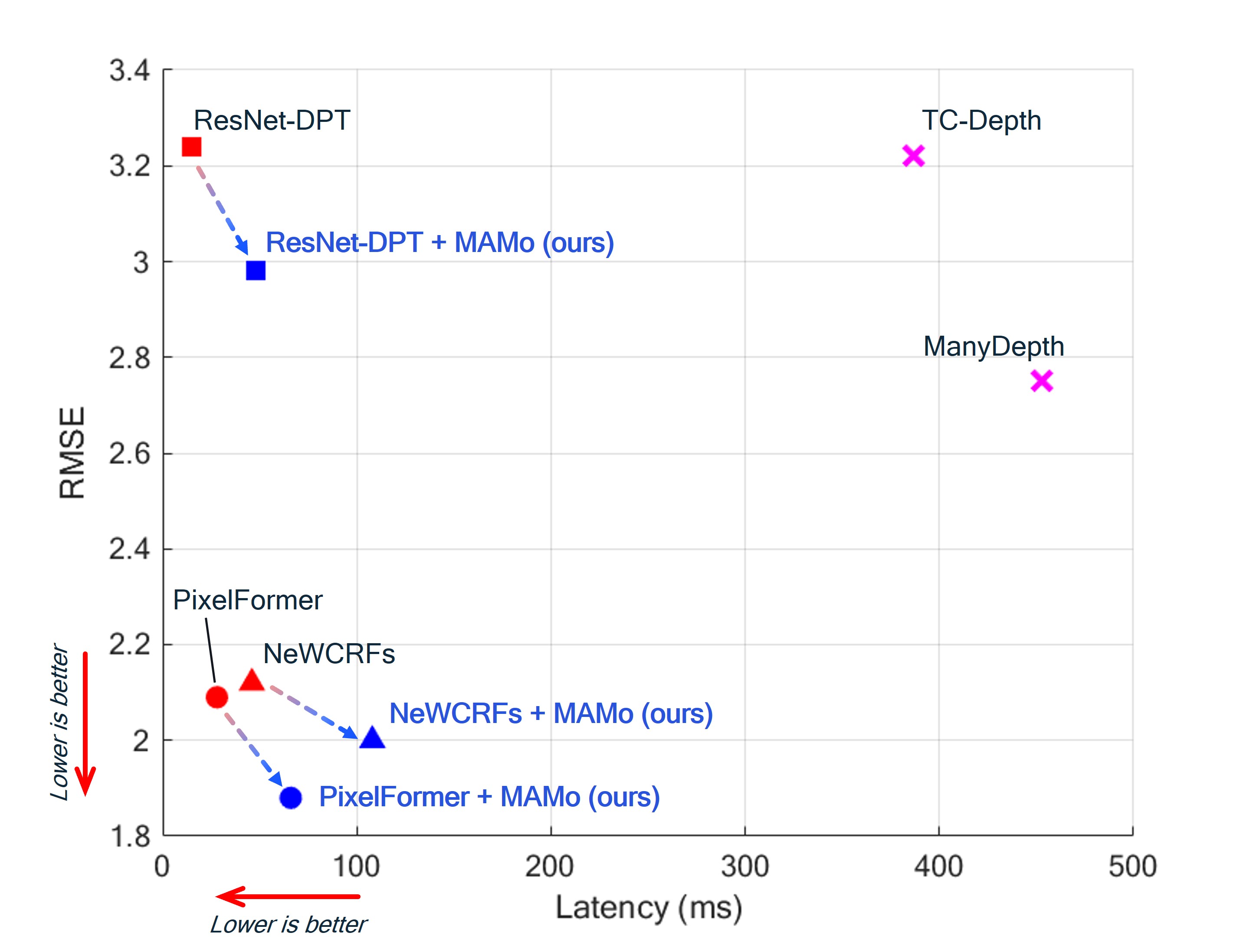} 
    \vspace{-2pt}
    \caption{RMSE vs. Latency on KITTI Eigen test set.}
    \label{fig:latency}
    \vspace{-5pt}
\end{figure}

\subsection{Computation: Accuracy vs. Efficiency}\vspace{-4pt}
We compare the average inference time using our proposed MAMo models with existing state-of-the-art multi-frame video depth estimation methods on KITTI. In Fig.~\ref{fig:latency}, we see that MAMo enables considerable accuracy improvements while incurring minor additional runtime. On the other hand, ManyDepth and TC-Depth have significantly larger latencies. For instance, ManyDepth has a latency of over 450 ms while PixelFormer + MAMo requires 66 ms, and PixelFormer + MAMo provides significantly better accuracy as compared to ManyDepth.

Although we do not include the optical flow inference times in the figure, the latency of modern optical flow models are not large. For instance, RAFT and RAFT-small have latencies of 115 ms and 60 ms on KITTI images, respectively. As such, even in the case where the optical flow and depth models are run in a sequential manner, the MAMo-based models still have significantly lower latencies as compared to ManyDepth and TC-Depth. Latencies are measured using a 11GB NVIDIA RTX-2080 GPU.



\subsection{Qualitative Results on KITTI} \vspace{-4pt}
Fig.~\ref{fig:vissualizations} shows that MAMo considerably improves depth estimation over baselines PixelFormer and NeWCRFs in several regions, e.g., the traffic sign and the building facade in the left sample, the biker in the middle sample, and the fences in the right sample (highlighted by white boxes). Overall, MAMo provides clearer and sharper depth maps.

\vspace{-1pt}
\section{Conclusions}\vspace{-5pt}
In this paper, we proposed a novel monocular video depth estimation approach, MAMo, which leverages memory and attention, and can be applied to any existing monocular depth estimation networks to transform them into video prediction models. Specifically, in MAMo, we propose a novel memory update method that allows the memory to maintain relevant and useful information for depth estimation, as the model goes through a video. We device attention schemes to combine the information from the memory with the visual features of the current time, in order to predict accurate depth for the current input frame. 
Our extensive results and ablation study on: KITTI, NYU Depth V2, and DDAD, confirms that our MAMo approach is effective, improving monocular depth estimation accuracy consistently.

\newpage
\appendix
\twocolumn[{%
 \centering
 \LARGE Supplementary for \titlecolor{MAMo:} Leveraging \titlecolor{\underline{\textcolor{black}{M}}}emory and \titlecolor{\underline{\textcolor{black}{A}}}ttention for \titlecolor{\underline{\textcolor{black}{Mo}}}nocular Video Depth Estimation \vspace{-10pt}\\[1.5em]
}]
\section{Architecture Details}\vspace{-3pt}
In this section we explain in more detail how we apply MAMo to the latest SOTA monocular depth estimation methods to perform video depth estimation, including PixelFormer~\cite{Agarwal_2023_WACV}, NeWCRFs~\cite{yuan2022newcrfs}, and a strong convolutional baseline which is a variant of DPT~\cite{ranftl2021vision} with a ResNet encoder (referred to as ResNet-DPT).

\begin{figure*}[t!]
    \centering
    \includegraphics[width=0.98\linewidth]{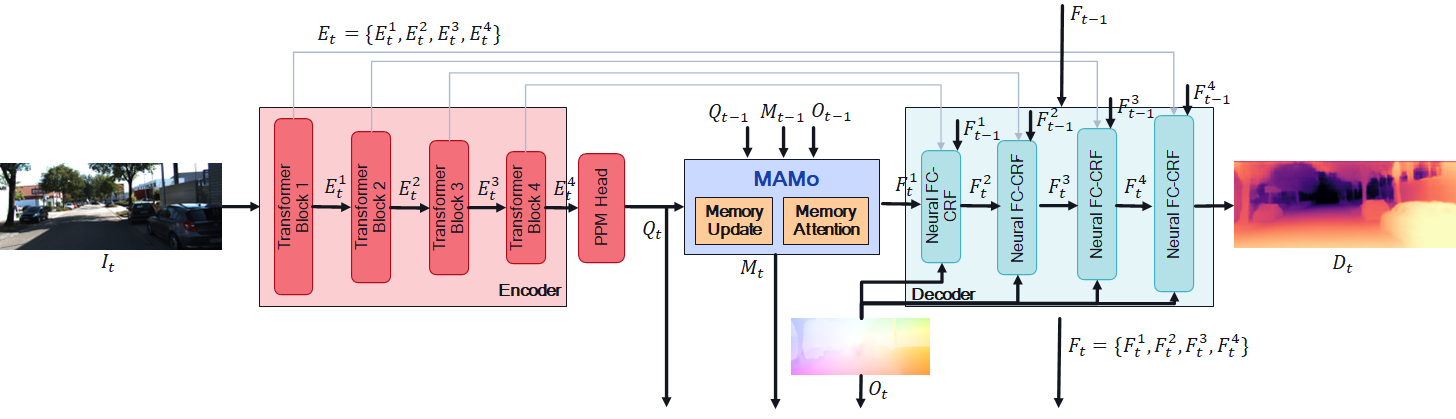} 
    \vskip -2pt 
    \caption{Detailed Architecture of NewCRFs + MAMo.}
    \label{fig:newcrfs+mamo}
    \vspace{-0.0em}
\end{figure*}

\begin{figure}[t!]
    \centering
    \includegraphics[width=0.98\linewidth]{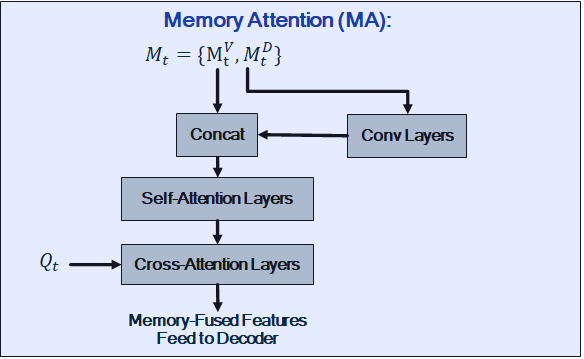} 
    \vskip -2pt 
    \caption{Overview of proposed Memory Attention in MAMo. For Self-attention and cross-attention, we use Neural FC-CRFs for NeWCRFs + MAMo, Skip Attention Module (SAM) for PixelFormer + MAMo, and LinFormer for ResNet-DPT + MAMo.}
    \label{fig:memory_attention}
    \vspace{-0.0em}
\end{figure}

\subsection{NeWCRFs + MAMo}\vspace{-3pt}
We apply our proposed MAMo approach to NeWCRFs~\cite{yuan2022newcrfs}, and refer to it as NeWCRFs + MAMo. We use follow same encoder and decoder architectures in ~\cite{yuan2022newcrfs}. For the encoder, Swin transformer~\cite{liu2021swin} is employed to extract the features. Pyramid Pooling Module~\cite{pintore2021slicenet} is used to extract global information. Pairwise potential module (PPM) head aggregates the global and local information. For the decoder, Neural Window FC-CRFs modules are employed to compute depth $D_t$.\footnote{See~\cite{yuan2022newcrfs} for more details on Neural Window FC-CRFs}. Since we concatenate optical flow $O_t$, the previous frame's decoder features $F_{t-1}$, and the current frame's encoder features $E_t$ as input to the decoder, we adjust the input channels of each Neural FC-CRF module of the decoder accordingly. Fig.~\ref{fig:newcrfs+mamo} shows a more detailed architectural view of NeWCRFs + MAMo. 

Fig.~\ref{fig:memory_attention} provides an illustration of the Memory Attention part in MAMo. For self-attention and cross-attention layers in NeWCRFs + MAMo, we use Neural Window FC-CRFs. 

\subsection{PixelFormer + MAMo}\vspace{-3pt}
We apply MAMo to PixelFormer~\cite{Agarwal_2023_WACV} and refer to it as PixelFormer + MAMo. We use the same architectures from~\cite{Agarwal_2023_WACV} for the encoder and decoder of PixelFormer + MAMo. For the encoder, Swin transformer~\cite{liu2021swin} is employed to extract the features. Pixel Query Initialise (PQI) is used to extract global information using pyramid spatial pooling~\cite{he2015spatial}, and compute the initial pixel queries $Q_t$. For the decoder, Skip Attention Modules (SAM) are employed to compute depth $D_t$.\footnote{See~\cite{Agarwal_2023_WACV} for more details on SAM.} The input channels of SAM modules are adjusted according to the concatenation of $E_t$, $F_{t-1}$ and $O_t$. We use SAM for the self-attention and cross-attention layers in the Memory Attention of PixelFormer + MAMo.

\subsection{ResNet-DPT + MAMo}\vspace{-3pt}
We apply MAMo to ResNet-DPT~\cite{ranftl2021vision}, and refer to it as ResNet-DPT + MAMo.  For the encoder, ResNet50~\cite{he2016deep} is employed to extract the features. For the decoder, we use the fusion module from~\cite{ranftl2021vision} to compute depth $D_t$. For self-attention and cross-attention layers in the Memory Attention of ResNet-DPT + MAMo, we use LinFormer attention modules~\cite{wang2020linformer}.

\section{Training Details}\vspace{-3pt}
Detailed training steps are provided in Algorithm~\ref{alg:train}. Note, we train the networks PixelFormer, NeWCRFs, and ResNet-DPT for first 5 epochs without MAMo, and train PixelFormer+MAMo, NeWCRFs+MAMo, and ResNet-DPT+MAMo with MAMo for the rest 20 epochs.

\begin{algorithm}[t]
\caption{Training MAMo video depth model}\label{alg:train}
\footnotesize{
   
\begin{algorithmic}[]
\STATE \textbf{Input}: Training dataset $\mathcal{D}_V$ consisting of training videos and depth ground truths. For each training video, $V =\{I_0,...,I_T\}$ and $D^{gt} =\{D^{gt}_0,...,D^{gt}_T\}$
\STATE \textbf{Model}: $h(\cdot)$ and $g(\cdot)$: encoder and 
 full depth network

\FOR{every epoch}
    \FOR {$V,D^{gt} \in \mathcal{D}_V$}
        \STATE \textit{\textbf{Initialization}}
        \STATE \quad $Q_0 \gets h(I_0),\ \ O_0 \gets \mathbf{0},\ \ F_{-1} \gets \mathbf{0}$
        \STATE \quad \textit{Update} $M_{0}$ \quad (repeat $Q_0 \ \ and \ O_0$ for \textit{L} times)
        \STATE \quad $D_0 \gets g(I_0; M_0, O_0, F_{-1})$
        \FOR {$I_t,D^{gt}_t \in V,D^{gt}$}
            \STATE \quad \textit{Estimate} $O_t$
            \STATE \textit{\textbf{Memory Update}} \ \ (Sec. 3.2 in the main paper)
            \STATE \quad $\widetilde{M}_t^V \gets \{M_{t-1}^V, Q_{t-1}\}, \ \ \widetilde{M}_t^D \gets \{M_{t-1}^D, O_{t-1}\}$ 
            \STATE \quad $\widetilde{M}_t \gets \{\widetilde{M}_t^V, \widetilde{M}_t^D\}$
            \STATE \quad $I_t^w \gets \textit{Warp}(I_{t-1}, O_t)$
            \STATE \quad $\widetilde{D}_{t} \gets g(I_t;\widetilde{M}_t, O_t, F_{t-1})$
            \STATE \quad  $\widetilde{D}_t^w \gets g(I_t^w;\widetilde{M}_t, O_t, F_{t-1})$
            \STATE \quad \textit{SILogLoss} ($\widetilde{D}_{t}$, $\widetilde{D}_t^w$)
            \STATE \quad \textit{Backpropagation}
            \STATE \quad \textit{Update} $M_t$ \ \ (Eq. 2 in the main paper)

            \STATE \textit{\textbf{Depth Estimation}}
            \STATE \quad $D_t \gets g(I_t; M_t, O_t, F_{t-1})$,\quad $Q_t \gets h(I_t)$
            \STATE \quad \textit{Compute} $\mathcal{L}_d$ \textit{between} $D_t$ \textit{and} $D^{gt}_t$ 
            \STATE \quad \quad \quad \quad \quad  \quad \quad \quad \quad \quad \quad \quad \quad (Eq. 5 in the main paper) 
            \STATE \quad \textit{Update parameters of} $h(\cdot)$, $g(\cdot)$

        \ENDFOR
    \ENDFOR
\ENDFOR
\end{algorithmic}
}
\end{algorithm}

\begin{table*}[t!]
    \caption{Quantitative results on KITTI (Eigen split) for distances up to 80 meters. $\dagger$ means methods uses multiple networks to estimate depth. ManyDepth-FS, and TC-Depth-FS means ManyDepth and TC-Depth are trained in fully-supervised fashion using ground-truths respectively. MF means multi frame methods, SF means single frame methods, and VD means extending MDE to VDE methods.$\uparrow$ means higher the better, and $\downarrow$ means lower the better. }
	\label{tab:KITTI}
    \centering
    \resizebox{1\linewidth}{!}{
    \normalsize
    \begin{tabular}{ll|l|ccccccc}
    \hline
        Type & Method & Encoder & Abs Rel$\downarrow$  & Sq Rel$\downarrow$ & RMSE$\downarrow$ & $\text{RMSE}_{log}\downarrow$ & $\delta<1.25\uparrow$ & $\delta<1.25^2\uparrow$ & $\delta<1.25^3\uparrow$  \\ \hline
        \multirow{12}{*}{MF} & NeuralRGB~\cite{liu2019neural} &  CNN based$\dagger$ & 0.100 & -- & 2.829 & --  & 0.931 & -- & --\\ 
        & ST-CLSTM~\cite{zhang2019exploiting} & Resnet18 & 0.101 & -- & 4.137 & --  & 0.890 & 0.970 & 0.9890\\ 
        & FlowGRU~\cite{eom2019temporally} & CNN~\cite{eom2019temporally} & 0.112 & 0.700 & 4.260 & 0.184  & 0.881 & 0.962 & 0.9830\\ 
        & Flow2Depth~\cite{xie2020video} &  CNN~\cite{mayer2016large}$\dagger$ & 0.081 & 0.488 & 3.651 & 0.146  & 0.912 & 0.970 & 0.9883\\ 
        & RDE-MV~\cite{patil2020don} &  ResNet18$\dagger$ & 0.111 & 0.821 & 4.650 & 0.187  & 0.821 & 0.961 & 0.9823\\ 
        & Patil \textit{et.al.}~\cite{patil2020don} &  ResNet18$\dagger$+ConvLSTM & 0.102 & -- & 4.148 & --  & 0.884 & 0.961 & 0.9824\\
        & Cao \textit{et.al.}~\cite{cao2021learning} &  -- & 0.099 & -- & 3.832 & --  & 0.886 & 0.968 & 0.9890\\ 
        & STAD~\cite{lee2021stad} & CNN $\dagger$~\cite{liu2019neural} & 0.109 & 0.594 & 3.312 & 0.153  & 0.889 & 0.971 & 0.9890\\ 
        & FMNet~\cite{wang2022less} &  ResNeXt-101 & 0.099 & -- & 3.832 & 0.129  & 0.886 & 0.968 & 0.9893\\  
        & ManyDepth-FS~\cite{watson2021temporal} &  ResNet50 & 0.069 & 0.342 & 3.414 & 0.111  & 0.930 & 0.989 & 0.9970\\
        & ManyDepth-FS~\cite{watson2021temporal} &  Swin-large & 0.060 & 0.248 & 2.747 & 0.099  & 0.955 & 0.993 & 0.9981\\
        & TC-Depth-FS~\cite{ruhkamp2021attention} &  ResNet50 & 0.071 & 0.330 & 3.222 & 0.108  & 0.922 & 0.993 & 0.9970\\\hline
        \multirow{5}{*}{SF} & AdaBins~\cite{bhat2021adabins} &  EfficientNet-B5+mViT~\cite{tan2019efficientnet} &0.058	& 0.190	& 2.360	&0.088	&0.964	&0.995	&0.9991\\
        & BinsFormer~\cite{li2022binsformer} & Swin-large &0.052	&0.151	&2.098	&0.079&0.975	&0.997	&0.9992	\\	
        & DepthFormer~\cite{agarwal2022depthformer} & MiT-B4~\cite{xie2021segformer} &0.058	&0.187	& 2.285	&0.087	&0.967	&0.996	&0.9991\\
        
        & SwinV2-MIM~\cite{xie2023darkmim} & Swin-large & 0.050 & 0.139 & 1.966 &	0.075&	0.977&	0.998&	\textbf{0.9995}	\\
        & URCDC~\cite{shao2023urcdc} & Swin-large & 0.050 & 0.142 & 2.032 & 0.076 & 0.977 & 0.997 & 0.9994 \\ 
        \hline
        \multirow{8}{*}{VD} &  ResNet-DPT &  ResNet50 & 0.085 & 0.383 & 3.242 & 0.130 & 0.913 & 0.981  & 0.9960 \\
        & \cellcolor{lightgray} ResNet-DPT+MAMo (ours) & \cellcolor{lightgray} ResNet50 & \cellcolor{lightgray} 0.071 & \cellcolor{lightgray} 0.301 & \cellcolor{lightgray} 2.984 & \cellcolor{lightgray} 0.121  & \cellcolor{lightgray} 0.926 & \cellcolor{lightgray} 0.990  & \cellcolor{lightgray} 0.9971  \\
        & NeWCRFs~\cite{yuan2022newcrfs} & Swin-Base & 0.054 & 0.157 & 2.140 & 0.081 & 0.973 & 0.997 & 0.9993  \\ 
        & \cellcolor{lightgray} NeWCRFs+MAMo (ours)  & \cellcolor{lightgray} Swin-Base & \cellcolor{lightgray} 0.051 & \cellcolor{lightgray} 0.149 & \cellcolor{lightgray} 2.090 & \cellcolor{lightgray} 0.078 & \cellcolor{lightgray} 0.976 & \cellcolor{lightgray} 0.998 & \cellcolor{lightgray} 0.9994  \\ 
        & NeWCRFs & Swin-large & 0.053 & 0.154 & 2.118 & 0.080 & 0.974 & 0.997 & 0.9994  \\
        & \cellcolor{lightgray} NeWCRFs+MAMo (ours) & \cellcolor{lightgray} Swin-large & \cellcolor{lightgray} 0.050 & \cellcolor{lightgray} 0.141 & \cellcolor{lightgray} 2.003 & \cellcolor{lightgray} 0.076 & \cellcolor{lightgray} \textbf{0.977} & \cellcolor{lightgray} \textbf{0.998} & \cellcolor{lightgray} 0.9994  \\ 
        & PixelFormer~\cite{Agarwal_2023_WACV} & Swin-large & 0.052 & 0.152 & 2.093 & 0.079 & 0.975 & 0.997 & 0.9994  \\
        & \cellcolor{lightgray} PixelFormer+MAMo (ours) & \cellcolor{lightgray} Swin-large & \cellcolor{lightgray} \textbf{0.049} & \cellcolor{lightgray} \textbf{0.130} & \cellcolor{lightgray} \textbf{1.884} & \cellcolor{lightgray} \textbf{0.072} & \cellcolor{lightgray} \textbf{0.977} & \cellcolor{lightgray} \textbf{0.998} & \cellcolor{lightgray} \textbf{0.9995}  \\ 
         \hline
    \end{tabular}
    }
\end{table*}
\begin{table}[t!]
    \caption{Quantitative results on DDAD dataset for distances up to 200 meters, and input frame resolution is $1216\times 1936$.}
	\label{tab:DDAD}
    \centering
    \resizebox{1\linewidth}{!}{
    \small
    \begin{tabular}{l|l|ccc}
    \hline
        Method & Encoder & Sq Rel$\downarrow$ & RMSE$\downarrow$ & $\delta<1.25\uparrow$ \\ \hline
        ManyDepth-FS~\cite{watson2021temporal} & Swin-large & 4.211 & 13.899 & 0.784 \\ 
        SwinV2-MIM\cite{xie2023darkmim} & Swin-large & 3.505 & 11.641 & 0.853  \\ 
        \hline
        NeWCRFs & Swin-large & 4.041 & 11.956 & 0.816 \\ 
        \rowcolor{lightgray} NeWCRFs+MAMo (ours) & Swin-large & \textbf{2.990} & \textbf{10.462} & 0.867 \\  
        PixelFormer &  Swin-large & 4.474& 12.467 & 0.802 \\ 
        \rowcolor{lightgray} PixelFormer+MAMo (ours) & Swin-large & 3.349 & 11.094 & \textbf{0.870}    \\ 
        \hline
    \end{tabular}
    }
    \vspace{-0.0em}
\end{table}
\subsection{Temporal consistency} 
We evaluate temporal consistency using the metrics from Li et al.~\cite{li2021enforcing},
\begin{equation}
\nonumber
\resizebox{\linewidth}{!}{$
\begin{aligned}
    aTC_t &= \frac{1}{\sum(K_t==1)}K_t\|\frac{D_t - {D}_{t}^w}{D_t}\|,\\
    rTC_t &=  \frac{1}{\sum(K_t==1)}K_t\left[\text{Max}\left(\frac{D_t}{{D}_{t}^w}, \frac{{D}_{t}^w}{K_t}  \right)<\text{thr}\right],
\end{aligned}
$} 
\end{equation}
where $K_t$ is a depth validity mask, $D_t$ is predicted depth for $I_t$ and ${D}_{t}^w$ is warped from ${D}_{t-1}$ using optical flow; we use the latest SOTA FlowFormer~\cite{huang2022flowformer}.
Table~\ref{tab:TC_comp} shows that MAMo is more temporally consistency than both the monocular baseline, as well as SOTA ManyDepth and TC-Depth. 

\begin{table}[t!]
    \caption{\small Temporal consistency evaluation on KITTI. We use Swin-Large encoder for NeWCRFs and NeWCRFs + MAMo.}
	\label{tab:TC_comp}
    \centering
    \resizebox{0.98\linewidth}{!}{
    \normalsize
    \begin{tabular}{c|ccc|ccc}
    \hline
        \multirow{2}{*}{Metrics} & \multirow{2}{*}{ManyDepth} & \multirow{2}{*}{TC-Depth} & \multirow{2}{*}{NeWCRFs} &\multicolumn{3}{c}{NeWCRFs + MAMo}  \\ 
        &              & &      &            L=2 &            L=4        &           L=6          \\ \hline
         rTC $\uparrow$ & 0.920 & 0.901 & 0.914 &   0.952  &   0.963  & 0.966   \\ 
         aTC $\downarrow$ & 0.111 & 0.122 &  0.116 &   0.091  &    0.088 & 0.086\\ 
        \hline
    \end{tabular}
    }
\end{table}

\section{Additional Results}\vspace{-3pt}
In this section, we provide additional comparison results with latest, unpublished methods, as well as additional ablation studies.

\subsection{Additional Comparison on KITTI and DDAD}\vspace{-3pt}
In Table~\ref{tab:KITTI}, we provide a more comprehensive comparison that includes latest unpublished methods, such as Swin-MIM~\cite{xie2023darkmim} and and URCDC~\cite{shao2023urcdc} on KITTI.

In Table~\ref{tab:DDAD}, we further include Swin-MIM~\cite{xie2023darkmim} in the comparison on DDAD, where the models are trained on KITTI and tested on DDAD.

\subsection{Additional Ablation Studies}\vspace{-3pt}
\subsubsection{Token Channels}\vspace{-3pt}
We perform an ablation study for different number of feature channels in the visual memory tokens. As shown in Table~\ref{tab:Abl_token}, when using NeWCRFs + MAMo, the model's accuracy is almost the same for token channels of 256 and 512 (we use 512 in the main paper). This allows one to improve computational efficiency as needed with slight accuracy drops. 


\subsubsection{Augmentation of Frame Subsampling}\vspace{-3pt}
In the paper, we use frame subsampling as an augmentation when training the video depth model (c.f.~Section 3.5 in the main paper).  Table~\ref{tab:Abl_droprate} provides an ablation study for not using and using frame subsampling, with drop rates $r$ equal to 0 and 4, respectively. It can be seen that frame subsampling leads to lower depth estimation errors, since it allows the network to see more variety of motion and scene changes.

\subsection{Qualitative Results}
\begin{figure*}[t]
    \centering
    \includegraphics[width=0.6\linewidth]{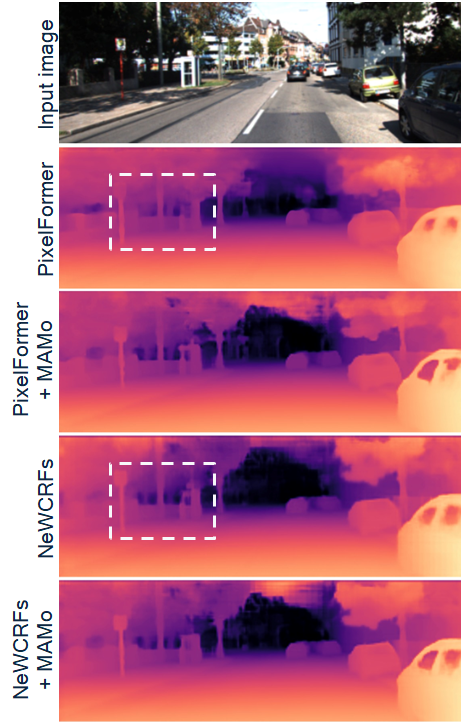}
    \caption{Qualitative results on KITTI. We highlight (white boxes) regions where MAMo significantly improves depth estimation quality.}
    \vspace{-1pt}
    \label{fig:vissualizations_1}
    \vspace{-0.0em}
\end{figure*}

\begin{figure*}[t]
    \centering
    \includegraphics[width=0.8\linewidth]{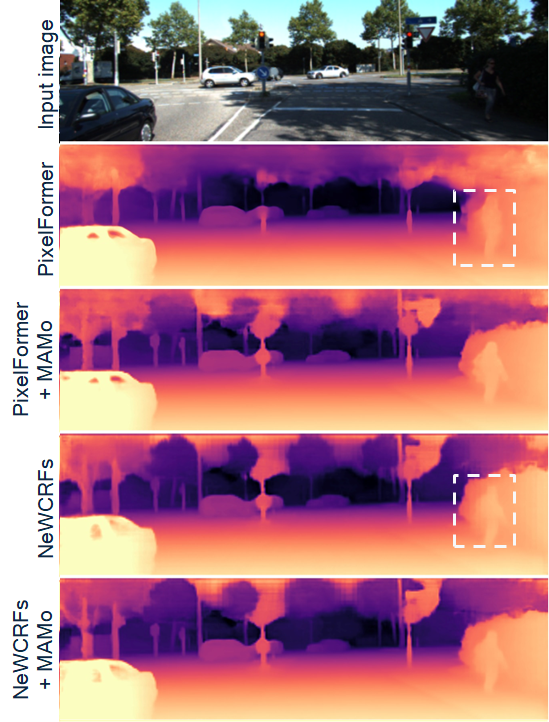}
    \caption{Qualitative results on KITTI. We highlight (white boxes) regions where MAMo significantly improves depth estimation quality.}
    \vspace{-1pt}
    \label{fig:vissualizations_2}
    \vspace{-0.0em}
\end{figure*}

\begin{figure*}[t]
    \centering
    \includegraphics[width=0.8\linewidth]{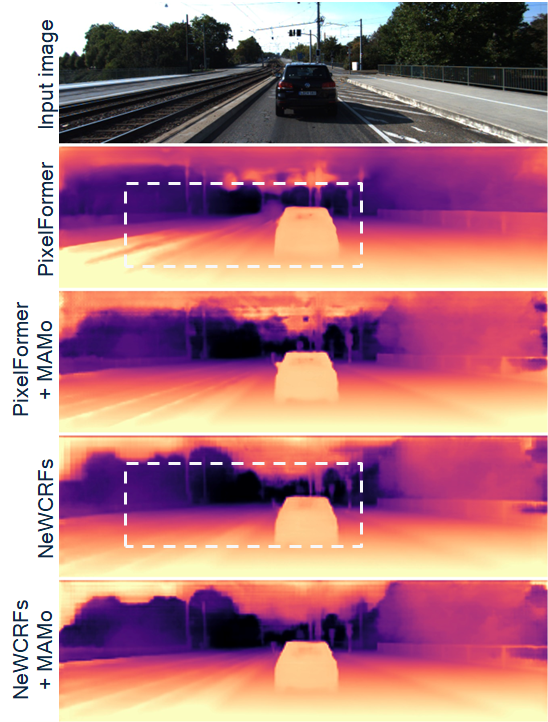}
    \caption{Qualitative results on KITTI. We highlight (white boxes) regions where MAMo significantly improves depth estimation quality.}
    \vspace{-1pt}
    \label{fig:vissualizations_3}
    \vspace{-0.0em}
\end{figure*}

\vspace{-3pt}
We provide additional visual results. Figures~\ref{fig:vissualizations_1},~\ref{fig:vissualizations_2},~and~\ref{fig:vissualizations_3} show that MAMo considerably improves depth estimation over baselines PixelFormer and NeWCRFs in several regions: (i) traffic sign and telephone booth in Fig.~\ref{fig:vissualizations_1}, (ii) person in Fig.~\ref{fig:vissualizations_2}, and (iii) railway tracks and car in Fig.~\ref{fig:vissualizations_3}.

\section{Optical Flow Estimation Models}

We use the official codes and pre-trained checkpoints from RAFT.\footnote{\url{https://github.com/princeton-vl/RAFT}}
We use Sintel-trained checkpoint for indoor scenarios like NYU-Depth V2 and KITTI-trained checkpoint for outdoor scenarios like KITTI and DDAD. 
\begin{table}[t!]
    \caption{Ablation experiment for number of channels in visual memory token on KITTI dataset. We perform this experiment using NeWCRFs + MAMo with Swin-Large encoder. }
	\label{tab:Abl_token}
    \centering
    \resizebox{1\linewidth}{!}{
    \normalsize
    \begin{tabular}{c|ccccc}
    \hline
        \begin{tabular}[c]{@{}c@{}}Token \\Channels\end{tabular}& Abs Rel$\downarrow$  & Sq Rel$\downarrow$ & RMSE$\downarrow$  & $\delta<1.25\uparrow$ & $\delta<1.25^2\uparrow$ \\ \hline
         $256$ &  0.050 & 0.140 &  2.025  & 0.977 & 0.998    \\ 
         $512$  & 0.050  & 0.141 & 2.003  & 0.977   &  0.998 \\ 
        \hline
    \end{tabular}
    }
    \vspace{-0em}
\end{table}

\begin{table}[t!]
    \caption{Ablation experiment for Frame sampling on KITTI dataset. We perform this experiment using NeWCRFs + MAMo with Swin-Large encoder.}
	\label{tab:Abl_droprate}
    \centering
    \resizebox{1\linewidth}{!}{
    \normalsize
    \begin{tabular}{c|ccccc}
    \hline
        Drop Rate & Abs Rel$\downarrow$  & Sq Rel$\downarrow$ & RMSE$\downarrow$  & $\delta<1.25\uparrow$ & $\delta<1.25^2\uparrow$ \\ \hline
         r = 0 &  0.050 & 0.142  &  2.032 & 0.977   &  0.998    \\ 
         r = 4 & 0.050  & 0.141 & 2.003  & 0.977   &  0.998  \\ 
        \hline
    \end{tabular}
    }
    \vspace{-0em}
\end{table}

\setlength{\textfloatsep}{10pt} 
\section{NYUDv2-Video}
\subsection{NYUDv2-Video: Training set}
Image frames from following scenes of NYU Depth v2 dataset are used as training images for NYUDv2-Video:
\\
basement\_0001a,
basement\_0001b,
bathroom\_0001,\\
bathroom\_0002,
bathroom\_0005,
bathroom\_0006,\\
bathroom\_0007,
bathroom\_0010,
bathroom\_0011,\\
bathroom\_0014a,
bathroom\_0016,
bathroom\_0019,\\
bathroom\_0023,
bathroom\_0024,
bathroom\_0028,\\
bathroom\_0030,
bathroom\_0034,
bathroom\_0035,\\
bathroom\_0039,
bathroom\_0041,
bathroom\_0042,\\
bathroom\_0045a,
bathroom\_0048,
bathroom\_0049,\\
bathroom\_0050,
bathroom\_0056,
bathroom\_0057,\\
bedroom\_0004,
bedroom\_0012,
bedroom\_0015,\\
bedroom\_0017,
bedroom\_0019,
bedroom\_0021,\\
bedroom\_0025,
bedroom\_0028,
bedroom\_0029,\\
bedroom\_0033,
bedroom\_0034,
bedroom\_0035,\\
bedroom\_0036,
bedroom\_0039,
bedroom\_0040,\\
bedroom\_0041,
bedroom\_0042,
bedroom\_0045,\\
bedroom\_0047,
bedroom\_0050,
bedroom\_0051,\\
bedroom\_0052,
bedroom\_0056a,
bedroom\_0056b,\\
bedroom\_0057,
bedroom\_0060,
bedroom\_0062,\\
bedroom\_0065,
bedroom\_0067a,
bedroom\_0067b,\\
bedroom\_0071,
bedroom\_0076a,
bedroom\_0078,\\
bedroom\_0079,
bedroom\_0080,
bedroom\_0081,\\
bedroom\_0082,
bedroom\_0086,
bedroom\_0094,\\
bedroom\_0097,
bedroom\_0098,
bedroom\_0100,\\
bedroom\_0104,
bedroom\_0107,
bedroom\_0118,\\
bedroom\_0120,
bedroom\_0124,
bedroom\_0125a,\\
bedroom\_0130,
bedroom\_0136,
bedroom\_0140,\\
bookstore\_0001d,
bookstore\_0001e,
bookstore\_0001f,\\
bookstore\_0001i,
bookstore\_0001j,
cafe\_0001a,\\
cafe\_0001b,
cafe\_0001c,
classroom\_0003,\\
classroom\_0004,
classroom\_0005,
classroom\_0006,\\
classroom\_0011,
classroom\_0016,
classroom\_0018,\\
computer\_lab\_0002,
conference\_room\_0001,\\
conference\_room\_0002,
dinette\_0001,\\
dining\_room\_0004,
dining\_room\_0008,\\
dining\_room\_0010,
dining\_room\_0012,
dining\_room\_0013,\\
dining\_room\_0014,
dining\_room\_0016,
dining\_room\_0024,\\
dining\_room\_0028,
dining\_room\_0031,
dining\_room\_0033,\\
dining\_room\_0034,
excercise\_room\_0001,
foyer\_0002,\\
furniture\_store\_0001a,
furniture\_store\_0001c,\\
furniture\_store\_0001d,
furniture\_store\_0001f,\\
furniture\_store\_0002b,
furniture\_store\_0002c,\\
furniture\_store\_0002d,
home\_office\_0004,
home\_office\_0005,
home\_office\_0006,
home\_office\_0008,
home\_office\_0011,
home\_office\_0013,
home\_storage\_0001,
indoor\_balcony\_0001,
kitchen\_0006,
kitchen\_0008,
kitchen\_0010,
kitchen\_0011b,
kitchen\_0016,
kitchen\_0028a,
kitchen\_0028b,
kitchen\_0029a,
kitchen\_0033,
kitchen\_0035a,
kitchen\_0037,
kitchen\_0043,
kitchen\_0045a,
kitchen\_0045b,
kitchen\_0049,
kitchen\_0051,
kitchen\_0052,
kitchen\_0053,
kitchen\_0059,
kitchen\_0060,
laundry\_room\_0001,\\
living\_room\_0005,
living\_room\_0010,
living\_room\_0012,
living\_room\_0020,
living\_room\_0022,
living\_room\_0032,
living\_room\_0033,
living\_room\_0035,
living\_room\_0037,
living\_room\_0038,
living\_room\_0040,
living\_room\_0042a,
living\_room\_0046a,
living\_room\_0047b,
living\_room\_0055,
living\_room\_0058,
living\_room\_0063,
living\_room\_0068,
living\_room\_0069a,
living\_room\_0070,
living\_room\_0071,
living\_room\_0082,
living\_room\_0083,
living\_room\_0085,
living\_room\_0086a,
nyu\_office\_0,
nyu\_office\_1,\\
office\_0003,
office\_0004,
office\_0009,
office\_0012,
office\_0019,
office\_0021,
office\_0023,
office\_0024,
office\_0025,
office\_0026,
office\_kitchen\_0003,\\
playroom\_0002,
playroom\_0003,
playroom\_0004,\\
playroom\_0006,
printer\_room\_0001,\\
reception\_room\_0001a,
reception\_room\_0001b,\\
reception\_room\_0002,
reception\_room\_0004,\\
student\_lounge\_0001,
study\_0003,
study\_0004,
study\_0005,
study\_0006,
study\_0008,
study\_room\_0004,
study\_room\_0005a,
study\_room\_0005b,

\subsection{NYUDv2-Video: Test set}
Image frames from following scenes of NYU Depth v2 dataset are used as test images for NYUDv2-Video:

bathroom\_0013, 
bathroom\_0033, 
bathroom\_0051, 
bathroom\_0053, 
bathroom\_0054, 
bathroom\_0055, 
bedroom\_0010, 
bedroom\_0014, 
bedroom\_0016, \\
bedroom\_0020, 
bedroom\_0026, 
bedroom\_0031, \\
bedroom\_0038, 
bedroom\_0053, 
bedroom\_0059, \\
bedroom\_0063, 
bedroom\_0066, 
bedroom\_0069, \\
bedroom\_0072, 
bedroom\_0074, 
bedroom\_0090, \\
bedroom\_0096, 
bedroom\_0106, 
bedroom\_0113, \\
bedroom\_0116, 
bedroom\_0125b, 
bedroom\_0126, \\
bedroom\_0129, 
bedroom\_0132, 
bedroom\_0138, \\
bookstore\_0001g, 
bookstore\_0001h, 
classroom\_0010, \\
classroom\_0012, 
classroom\_0022, 
dining\_room\_0001b, \\
dining\_room\_0002, 
dining\_room\_0007, 
dining\_room\_0015, 
dining\_room\_0019, 
dining\_room\_0023, 
dining\_room\_0029, 
dining\_room\_0037, 
furniture\_store\_0001b, \\
furniture\_store\_0001e, 
furniture\_store\_0002a, 
home\_office\_0007, 
kitchen\_0003, 
kitchen\_0011a, 
kitchen\_0017, 
kitchen\_0019a, 
kitchen\_0019b, 
kitchen\_0029b, 
kitchen\_0029c, 
kitchen\_0031, 
kitchen\_0035b, 
kitchen\_0041, 
kitchen\_0047, 
kitchen\_0048, 
kitchen\_0050, 
living\_room\_0004, 
living\_room\_0006, \\
living\_room\_0011, 
living\_room\_0018, 
living\_room\_0019, 
living\_room\_0029, 
living\_room\_0039, 
living\_room\_0042b, 
living\_room\_0046b, 
living\_room\_0047a, 
living\_room\_0050, 
living\_room\_0062, 
living\_room\_0067, 
living\_room\_0069b, 
living\_room\_0078, 
living\_room\_0086b, 
office\_0006, 
office\_0011, 
office\_0018, 
office\_kitchen\_0001a, \\
office\_kitchen\_0001b,

\newpage
{\small
\bibliographystyle{ieee_fullname}
\bibliography{egbib}
}

\end{document}